\documentclass[conference, compsoc]{IEEEtran}
\IEEEoverridecommandlockouts
\ifCLASSOPTIONcompsoc
  \usepackage[nocompress]{cite}
\else
  \usepackage{cite}
\fi
\ifCLASSINFOpdf
  \usepackage[pdftex]{graphicx}
\else
  \usepackage[dvips]{graphicx}
\fi
\ifCLASSOPTIONcompsoc
  \usepackage[caption = false, font = footnotesize, labelfont = sf, textfont = sf]{subfig}
\else
  \usepackage[caption = false, font = footnotesize]{subfig}
\fi
\usepackage{mathrsfs}
\usepackage{amsmath, amssymb, amsfonts}
\usepackage{algorithmic, algorithm}
\usepackage{makecell}

\def\QEDclosed{%
  \hfill\ensuremath{\mbox{\rule[0pt]{1.3ex}{1.3ex}}}}

\begin{document}

\newcommand{\mc}{\mathcal}
\newcommand{\ul}{\underline}
\newcommand{\cd}{\centerdot}
\newcommand{\mb}{\mathbf}
\newcommand{\tr}{\textrm}
\newcommand{\tb}{\textbf}
\newcommand{\mr}{\mathrm}
\newcommand{\mbb}{\mathbb}
\newcommand{\ds}{\displaystyle}
\newcommand{\NN}[1]{{\mathscr{N}}(#1)}
\newcommand{\la}{\langle\;}
\newcommand{\ra}{\;\rangle}
\newcommand{\lla}{\left\langle}
\newcommand{\rra}{\right\rangle}
\newcommand{\wh}{\widehat}
\newcommand{\QCM}[2]{\ul{M}_{\ul{#2}}(\mc{#1})}
\newcommand{\Tr}{Tr}

\newtheorem{definition}{Definition}
\newtheorem{theorem}{Theorem}
\newtheorem{lemma}[theorem]{Lemma}
\newtheorem{corollary}[theorem]{Corollary}
\newtheorem{proposition}[theorem]{Proposition}


%
\title{%
  A Quantum Tensor Network-Based Viewpoint for Modeling and Analysis \\
  of Time Series Data}


\author{%
  \IEEEauthorblockN{%
    Pragatheeswaran Vipulananthan, Kamal Premaratne, Dilip Sarkar, Manohar N. Murthi}
  \IEEEauthorblockA{%
    University of Miami \\
    Coral Gables, Florida 33146 \\
    Email: \{pxv245, kamal, sarkar, mmurthi\}@miami.edu}
  \thanks{%
    This work is based on research supported by the National Science Foundation (NSF) Grant \#2123635 and Department of Defense (DoD) Grant \#W911NF-24-2-0216.}
}


%


\maketitle

\begin{abstract}
Accurate uncertainty quantification is a critical challenge in machine learning. While neural networks are highly versatile and capable of learning complex patterns, they often lack interpretability due to their ``black box'' nature. On the other hand, probabilistic ``white box'' models, though interpretable, often suffer from a significant performance gap when compared to neural networks. To address this, we propose a novel quantum physics-based ``white box'' method that offers both accurate uncertainty quantification and enhanced interpretability. By mapping the kernel mean embedding (KME) of a time series data vector to a reproducing kernel Hilbert space (RKHS), we construct a tensor network-inspired 1D spin chain Hamiltonian, with the KME as one of its eigen-functions or eigen-modes. We then solve the associated Schr{\"o}dinger equation and apply perturbation theory to quantify uncertainty, thereby improving the interpretability of tasks performed with the quantum tensor network-based model. We demonstrate the effectiveness of this methodology, compared to state-of-the-art ``white box" models, in change point detection and time series clustering, providing insights into the uncertainties associated with decision-making throughout the process.
\end{abstract}



%
\IEEEpeerreviewmaketitle


\section{Introduction}


\textbf{Interpretability.} 
Setting aside the ongoing debate regarding a universal definition, interpretability which tells us how the model arrives at the decision is a more useful notion (than explainability) for a more transparent and accountable model \cite{Rudin2019NatureMI_StopExplainingBlackBoxMLModels, Linardatos2021Entropy_ExplainableAI}. It broadly refers to the extent to which one can comprehend the reasons behind a decision. Interpretability, which  constitutes a critical factor in understanding, accountability, trustworthiness, and refinement of machine learning (ML) algorithms remains a persistent challenge \cite{carvalho2019machine}. These difficulties tend to be amplified in powerful deep neural network (NN) models. Indeed, it is widely acknowledged that such models lack interpretability \cite{ran2023tensor} and the high degree of nonlinearity associated with them only exacerbates the problem. Interpretability is a requirement for assessing robustness of ML algorithms too. Even a well-trained deep NN model can be greatly disrupted by noise or deliberate attacks \cite{su2019one}. Currently, there is a lack of comprehensive theories to quantitatively assess the impact of different disturbances on predictions and the confidence one may place on these predictions.

On the other hand, probabilistic ML models \cite{ghahramani2015probabilistic} are often considered ``white box'' approaches because they are inherently interpretable in terms of statistical means. For instance, probabilistic reasoning can uncover hidden causal relationships \cite{tenenbaum2006theory, lucas2010learning}. However, despite their interpretability, there is a significant performance gap between these probabilistic models and state-of-the-art deep NNs.

While our work is applicable to a broader class of data, in this paper we restrict our attention to time series (TS) data. TS data occur across a wide variety of domains where they are used for forecasting, classification, clustering, motif identification, anomaly detection, and other purposes. Interpretability remains a daunting challenge for ML methods that deal with TS data too. While advancements utilizing long short-term memory (LSTM), gated recurrent units (GRUs), and temporal convolution networks (TCNs) are able to account for local statistical drifts or non-stationarity in TS data \cite{greff2016lstm, cho2014properties, bai2018empirical}, whether their ``black box'' viewpoint offers interpretability is debatable.


\textbf{Physics-Inspired Approaches.} 
Physics-inspired methods have recently emerged as a promising approach to address the lack of interpretability in ML models. These methods aim to provide a deeper understanding of the underlying patterns and relationships within complex datasets \cite{singh2021toward, perez2006matrix}. Broadly, they can be categorized into two types: physics-inspired NNs (PINNs) \cite{zhao2024adversarial} and physics-based models and systems \cite{singh2020time}  directly applied to solve specific ML problems. Though interpretability remains limited in PINNs due to their ``blackbox'' nature, physics-based models and systems offer a viable solution which balances interpretability with a modest compromise in performance. 

Quantum physics-inspired methods have recently emerged as a way to address lack of interpretability in ML models. In particular, based on the quantum information processing framework (QIPF) in \cite{Principe2010_Book}, the work in \cite{singh2020time, singh2021toward} first maps the data (which could be an image, a sentence, or a segment of a TS), typically in the form of a vector, to a \emph{reproducing kernel Hilbert space (RKHS)} determined a by a suitable positive semi-definite (p.s.d.) kernel \cite{Aronszajn1950ToAMS}. The \emph{kernel mean embedding (KME)} produced by this mapping can be viewed as an empirical probability density function (PDF) of the data \cite{Scholkopf2015SC_RandomVariablesViaRKHS}. It allows one to capture non-linear relationships in the data, learn complex patterns that may not be discernible in the original data space, and leverage the `kernel trick' to perform various statistical analyses directly in the data domain \cite{Muandet2017_Book}. By examining the transformed data representation in the form of the KME, one may gain a better understanding of the data distribution and the relationships between different data points. The authors of \cite{singh2020time, singh2021toward} use the KME to produce a `data' wave function associated with the time-independent Schr{\"o}dinger equation of a quantum harmonic oscillator (QHO). Given that the set of Hermite polynomials forms an orthonormal basis (ONB) for the eigen-modes of the Hamiltonian operator associated with this Schr{\"o}dinger equation \cite{Zwiebach2022_Book}, the KME-generated data wave function is then projected onto this set of Hermite polynomials. These projections are then perturbed and used as a proxy for `sensitivity analysis' thus yielding a versatile, deterministic, single-shot representation of data uncertainty that is more descriptive than simple moment-based metrics \cite{singh2020time, singh2021toward}. 


\textbf{Tensor Networks (TNs).} 
In contrast, spurred on by the work in \cite{chertkov2018computational, Qi2019Quantum}, we view the KME-generated data wave function as just \emph{one} eigen-mode solution of a Hamiltonian associated with a quantum tensor network (TN). 

TNs have demonstrated significant success as an efficient means of representing states of large-scale quantum systems, making them valuable for quantum physics-inspired ML \cite{orus2019tensor, cirac2021matrix, felser2021quantum}. TN models, which come in the form of matrix product states (MPS), tree TNs, and projected entangled pair states (PEPS), each with its own network structure \cite{perez2006matrix, oseledets2011tensor, shi2006classical, verstraete2006criticality}, provide a ``white box'' computational tool offering interpretability akin to classical probabilistic models \cite{cheng2018information}. In quantum physics-inspired ML, data are mapped to a Hilbert space, enabling alternative methods like quantum many-body feature mapping (QMFM) which employ Pauli operators to encode data samples directly into the Hilbert space of quantum many-body states \cite{biamonte2017quantum}. Despite progress, establishing a systematic framework for interpretability remains a challenge. This paper explores a TN-based quantum physics-inspired approach for TS data analysis, aiming to bridge the gap between efficiency and interpretability.


\textbf{Contributions.} 
Our work offers the following contributions:

(1)~We view the KME associated with a set of data points as a 1-D spin chain framework. This representation provides a physical interpretation of the otherwise abstract KME vector, with each element corresponding to a spin state of the chain. Leveraging this quantum physics-inspired framework offers several advantages. First, it allows quantum physics-inspired algorithms to be employed for tasks such as clustering, uncertainty quantification (UQ), and dimensionality reduction, potentially leading to more efficient and accurate algorithms for data analysis. Second, the spin chain representation facilitates the exploration of entanglement and correlations between different features encoded in the KME. This allows for a deeper understanding of the underlying data structure and the relationships between dimensions. 

(2)~We treat the 1-D spin chain representation as a quantum TN, a ``white box'' computational tool that offers interpretability, enabling the application of TN methods for efficient storage, computation, and visualization of high-dimensional data.

(3)~We also use the KME to generate a data wave function associated with the time-independent Schr{\"o}dinger equation, but in contrast to previous work  \cite{singh2020time, singh2021toward}, we view it as just \emph{one} eigen-mode solution of a Hamiltonian associated with a quantum TN. We then regard \emph{all} the eigen-mode solutions of this Hamiltonian---the wave function being one solution---as features of TS data for purposes of modeling and analysis. This method does not require us to project the data wave function onto any pre-specified ONB (viz., the set of Hermite polynomials associated with the QHO). 

(4)~Since we have access to the complete set of eigen-mode solutions of the spin Hamiltonian associated with the quantum TN, we are able to carry out a sensitivity analysis by applying perturbation theory, the mathematically sound well-established methodology from quantum physics for arriving at the eigen-modes and eigen-energies of the quantum TN with no recourse to the simpler QHO. 

Our approach, building on previous research (\cite{Principe2010_Book, singh2020time, singh2021toward, chertkov2018computational, Qi2019Quantum} offers a more interpretable computational tool for modeling and analysis of TS data. By utilizing ideas from perturbation theory, our framework can also be employed as a deterministic single-shot representation of the uncertainty associated with ML models.


\section{Preliminaries}


We use $\mbb{R}$ and $\mbb{C}$ to denote the reals and complex numbers, respectively; $\mbb{R}^N$ and $\mbb{R}^{M \times N}$ denote $N$-sized real-valued vectors and $(M \times N)$-sized real-valued matrices, respectively. $\mbb{C}^N$ and $\mbb{C}^{M \times N}$ are defined similarly for complex-valued entities. 

Given the $N$-length data vector
\begin{equation}
  \ul{x}
    = [x_1, x_2, \ldots, x_N]^T
    \in \mbb{R}^N,
\end{equation}
consider its Gaussian kernel-based empirical KME
\begin{equation}
  \psi^{\sharp} 
    = [\psi^{\sharp}_1, \ldots, \psi^{\sharp}_i, \ldots, \psi^{\sharp}_N]^T
    \in \mbb{R}^N,
\end{equation}
where
\begin{equation}
  \psi^{\sharp}_i
    = \sqrt{%
      \frac{1}{N}
       \sum_{j = 1}^N 
       \exp
       \left(
         -\frac{(x_i - x_j)^2}{2\sigma^2}
       \right)},\; 
       i = 1, \ldots, N.
\label{eq:KME}
\end{equation}


\subsection{1-D Spin System}


The computational basis associated with a 1-D spin system in quantum physics consists of vectors corresponding to different spin configurations. For a 1-D spin system with $L$ spins, these computational basis states are typically denoted using the ket notation $\mid s_1\, s_2\, \dots\, s_L \ra$, where $s_i,\; i = 1, \ldots, L$, represents either the spin-up operation $\mid\; \uparrow \ra$ or the spin-down operation $\mid\; \downarrow \ra$. 

For a 1-D spin system with only one spin, the basis states are $\mid\; \uparrow \ra$ and $\mid\; \downarrow \ra$. Hence, the KME $\ul{\psi}^{\sharp}$ of a one-spin 1-D spin system could be expressed as the linear combination
\begin{equation}
  \ul{\psi}^{\sharp} 
    = a\, \cdot \mid\; \uparrow \ra + b\, \cdot \mid\; \downarrow \ra,
\end{equation}
or more compactly, $\ul{\psi}^{\sharp} = [a, b]$. 

For a 1-D spin system with two spins, the basis states are $\mid\; \uparrow \uparrow \ra$, $\mid\; \uparrow \downarrow \ra$, $\mid\; \downarrow \uparrow \ra$, and $\mid\; \downarrow \downarrow \ra$. Hence, the KME $\ul{\psi}^{\sharp}$ of a two-spin 1-D spin system could be expressed as the linear combination
\begin{equation}
  \ul{\psi}^{\sharp} 
    = a \cdot | \uparrow \uparrow \ra + b \cdot | \uparrow \downarrow \ra + c \cdot | \downarrow \uparrow \ra + d \cdot | \downarrow \downarrow \ra,
\end{equation}
or more compactly, $\ul{\psi}^{\sharp} = [a, b, c, d]$. 

In general, to express a KME of length $N$, the number of spin particles we would need is $L = \log_2 N$, where, for simplicity, we assume that the value of $N$ is a power of $2$, viz,., $N = 2^L$. For instance, to represent a KME of length 256, 8 spin particles would be required. 


\section{Hamiltonian Target Space}
\label{sec:TargetSpace}


The Hamiltonian operator captures the total energy --- for our purposes the kinetic and potential energies --- of a quantum particle. The Schr{\"o}dinger equation describes the dynamic behavior of the total energy. The wave function constitutes a solution to the eigen-pair decomposition --- i.e., the eigen-functions or eigen-modes and eigen-energy values --- of the Hamiltonian operator. These eigen-pairs provide valuable insight into the system dynamics and contributes to model interpretability.

Given an arbitrary KME $\ul{\psi}^{\sharp}$, we aim to identify the set of Hamiltonians that have $\ul{\psi}^{\sharp}$ as one of its eigen-modes. In this paper, we restrict our attention to a 1-D spin chain of quantum particles that exhibit local nearest neighbor random couplings only. 

The Hermtian operators associated with a 1-D spin chain having one-site nearest neighbor random couplings take the form
\begin{multline}
  \sum_{\ell = 1}^L 
  c_{i0} 
  \cdot
  (\wh{\ul{\sigma}}_1^0 \otimes \dots \otimes \wh{\ul{\sigma}}_{\ell - 1}^0 \otimes \wh{\ul{\sigma}}_{\ell}^i \otimes \wh{\ul{\sigma}}_{\ell + 1}^0 \otimes \dots \otimes \wh{\ul{\sigma}}_L^0) \\
    = \sum_{\ell = 1}^L 
      c_{i0} 
      \cdot 
      (\wh{\ul{I}} \otimes \dots \otimes \wh{\ul{I}} \otimes \wh{\ul{\sigma}}_{\ell}^i \otimes \wh{\ul{I}} \otimes \dots \otimes \wh{\ul{I}}),
  \label{eq:OneSite}
\end{multline}
where $\wh{\ul{I}} \in \mbb{R}^{2 \times 2}$ is the identity operator, $\ell \in \{1, \ldots, L\}$ identifies the position of the operator on the spin chain, $i \in \{0, 1, 2, 3\}$ with 
\begin{alignat}{6}
  &\wh{\ul{\sigma}}_{\ell}^0 
    &
      &= \wh{\ul{I}} 
        &
          &= \begin{pmatrix}
               1 & 0 \\
               0 & 1
             \end{pmatrix};
            &\quad
              &\wh{\ul{\sigma}}_{\ell}^1 
                &
                  &= \wh{\ul{\sigma}}^x 
                    &
                      &= \begin{pmatrix}
                           0 & 1 \\
                           1 & 0
                         \end{pmatrix};
                         \notag \\
  &\wh{\ul{\sigma}}_{\ell}^2 
    &
      &= \wh{\ul{\sigma}}^y 
        &
          &= \begin{pmatrix}
               0 & -i \\
               i & 0
             \end{pmatrix};
            &\quad
              &\wh{\ul{\sigma}}_{\ell}^3 
                &
                  &= \wh{\ul{\sigma}}^z 
                    &
                      &= \begin{pmatrix}
                           1 & 0 \\
                           0 & -1
                         \end{pmatrix},
\end{alignat}
and $c_{ij} \in \mbb{R}$ are real-valued scalars. Here, $\wh{\ul{\sigma}}^x$, $\wh{\ul{\sigma}}^y$, and $\wh{\ul{\sigma}}^z$ are the Pauli operators in the $x$, $y$, and $z$ spin directions, respectively. With three spin directions to be accounted for, \eqref{eq:OneSite} yields 3 operators.  

When the spin chain has two-site nearest neighbor random couplings, these operators take the form
\begin{multline}
  \sum_{\ell = 1}^{L - 1} 
  c_{ij} 
  \cdot 
  (\wh{\ul{\sigma}}_1^0 \otimes \dots \otimes \wh{\ul{\sigma}}_{\ell}^i \otimes \wh{\ul{\sigma}}_{\ell + 1}^j \otimes \dots \otimes \wh{\ul{\sigma}}_L^0) \\
    = \sum_{\ell = 1}^{L - 1} 
      c_{ij} 
      \cdot 
      (\wh{\ul{I}} \otimes \dots \otimes \wh{\ul{\sigma}}_{\ell}^i \otimes \wh{\ul{\sigma}}_{\ell + 1}^j \otimes \dots \otimes \wh{\ul{I}}).
  \label{eq:TwoSite}
\end{multline}
With three spin directions to be accounted for, \eqref{eq:TwoSite} yields $3^2$ operators.  

In a similar manner, a spin chain having $K$-site nearest neighbor random couplings yield $3^K$ operators.

These nearest neighbor operators are all Hermitian and $\tr{span}\, (\mc{H})$, the span of the set $\mc{H}$ of these operators, constitutes the search space of our desired Hamiltonians. Note that the set $\mc{H}$ is pairwise orthonormal (w.r.t. the Hilbert-Schmidt inner product, or equivalently the Fr{\"o}benius inner product in finite dimensional linear spaces) \cite{angrisani2023learning}, i.e., $\forall\, \wh{\ul{H}}_i, \wh{\ul{H}}_j \in \mc{H}$, $\la \wh{\ul{H}}_j \mid \wh{\ul{H}}_i \ra = \Tr\, (\wh{\ul{H}}_i\, \wh{\ul{H}}_j^{\dag}) = \delta_{ij}$, where $\wh{(\cd)}^{\dag}$ denotes the adjoint operator and $\delta_{ij} = 1$ for $i = j$ and it is $0$ othewise. We note that each Hermitian operator in $\mc{H}$ is represented by a Hermitian matrix of size $(2^L \times 2^L) = (N \times N)$. So, $\mc{H} \subseteq \mbb{C}^{N \times N}$, the set of $(N \times N)$-sized complex-valued matrices.

In summary, our target space $\mc{H} = \{\wh{\ul{H}}_i,\; i \in 0, \ldots, T - 1\}$, where $T$ is a suitable positive integer, is finite and it is in $\tr{span}\, (\mc{H})$ that we seek a Hermitian operator that has the KME $\ul{\psi}^{\sharp} \in \mbb{R}^N$ as one of its eigen-modes.


\section{Quantum Correlation Matrix (QCM)}


Now that we have established the search space of Hamiltonians as $\tr{span}\, (\mc{H})$, we next need to identify those Hamiltonians which has the KME $\ul{\psi}^{\sharp}$ as an eigen-mode. We first need 

\begin{definition}[Quantum Correlation Matrix (QCM) \cite{Qi2019Quantum}]
\label{def:QCM}
The \emph{QCM} associated with the pairwise orthonormal target space of Hermitian operator set $\mc{H} = \{\wh{\ul{H}}_i,\; i \in 0, \ldots, T - 1\}$ w.r.t. to the vector $\ul{\psi}$ is $\QCM{H}{\psi} = \{(\QCM{H}{\psi})_{ij}\} \in \mbb{R}^{T \times T}$ whose $(i, j)$-th element is 
\[
  (\QCM{H}{\psi})_{ij}
    = 0.5\, \la \{\wh{\ul{H}}_i, \wh{\ul{H}}_j\} \ra_{\ul{\psi}} 
        - \la \wh{\ul{H}}_i \ra_{\ul{\psi}}
          \cdot
          \la \wh{\ul{H}}_j \ra_{\ul{\psi}}.  
\]
Here, $\la \cd \ra_{\ul{\psi}} \overset{\Delta}{=} \la \ul{\psi} \mid \cd \mid \ul{\psi} \ra$ is the expectation operator and $\{\wh{\ul{H}}_i,\, \wh{\ul{H}}_j\} = \wh{\ul{H}}_i \wh{\ul{H}}_j + \wh{\ul{H}}_j \wh{\ul{H}}_i$ denotes the anti-commutator of the operator pair $\{\wh{\ul{H}}_i, \wh{\ul{H}}_j\}$ in $\mc{H}$. 
\end{definition}

Note that $\QCM{H}{\psi}$ is real-valued, symmetric, and positive semi-definite (p.s.d.) \cite{Qi2019Quantum} (see Lemma~\ref{lem:M_realsymmetric} and Corollary~\ref{cor:M_NullSpace}). Thus, all its eigen-energies are non-negative real and they are associated with an orthonormal basis of eigen-modes which can be chosen as real-valued (see Corollary~\ref{cor:M_NullSpace}). Henceforth, we assume this to be the case. 

Next, let $\wh{\ul{H}} \in \tr{span}\, (\mc{H})$ so that it takes the form of the finite linear combination
\begin{equation}
  \wh{\ul{H}} 
    = \sum_{i = 0}^{T - 1}
      w_i \wh{\ul{H}}_i,\;
      \wh{\ul{H}}_i \in \mc{H},\;
      \ul{w} = \{w_i\} \in \mbb{R}^T.
  \label{eq:LinearCombination}
\end{equation}
Hereafter, for convenience, we will denote a linear combination of the type above as $\wh{\ul{H}} = (\mc{H})_{\ul{w}}$. Note that $\wh{\ul{H}}$ is a Hermitian operator as well; its variance w.r.t. the normalized vector $\ul{\psi},\; \la \ul{\psi} \mid \ul{\psi} \ra = 1$, is 
\begin{equation}
  \tr{Var}\, (\wh{\ul{H}})_{\ul{\psi}}
    = \la \wh{\ul{H}}^2 \ra_{\ul{\psi}} - |\la \wh{\ul{H}} \ra_{\ul{\psi}}|^2.
  \label{eq:Variance}
\end{equation}
Note that $\tr{Var}\, (\wh{\ul{H}})_{\ul{\psi}} \geq 0$ (see Lemma~\ref{lem:Var_positive}); $\tr{Var}\, (\wh{\ul{H}})_{\ul{\psi}} = 0$ $\iff$ $\ul{\psi}$ is an eigen-mode of $\wh{\ul{H}}$ \cite{Qi2019Quantum} (see Corollary~\ref{cor:H_eigen-state}). Thus, $\tr{Var}\, (\wh{\ul{H}})_{\ul{\psi}}$ can be viewed as a measure of the degree to which $\wh{\ul{H}}$ fails to have $\ul{\psi}$ as one of its eigen-modes.

More importantly, as pointed out in \cite{Qi2019Quantum}, the normalized vector $\ul{\psi},\; \la \ul{\psi} \mid \ul{\psi} \ra = 1$, is an eigen-mode of $\wh{\ul{H}} \in \tr{span}\, (\mc{H})$ iff the vector $\ul{w} \in \mbb{R}^T$ belongs to the null space of the QCM $\QCM{H}{\psi}$, i.e., $\ul{w} \in \NN{\QCM{H}{\psi}}$.

These observations allow us to formulate a strategy to identify those Hermitian operators in $\tr{span}\, (\mc{H})$ that has the KME $\ul{\psi}^{\sharp}$ as an eigen-mode.


\section{Spin Hamiltonian Candidates}


First, we make the following assumption: suppose there exists a large enough positive integer $T^{\sharp}$ s.t. $\ul{\psi}^{\sharp}$ is an eigen-mode of some $\wh{\ul{H}}^{\sharp} \in \tr{span}\, (\mc{H}^{\sharp})$, where $\mc{H}^{\sharp} = \{\wh{H}_i,\; i \in 0, \ldots, T^{\sharp} - 1\}$. If we now construct the QCM $\QCM{H^{\sharp}}{\psi^{\sharp}}$ associated with $\mc{H}^{\sharp}$ w.r.t. $\ul{\psi}^{\sharp}$ as (see Definition~\ref{def:QCM})
\begin{equation}
  (\QCM{H^{\sharp}}{\psi^{\sharp}})_{ij} 
    = 0.5\, \la \{\wh{\ul{H}}_i, \wh{\ul{H}}_j\} \ra_{\ul{v}} - \la \wh{\ul{H}}_i \ra_{\ul{v}} \cdot \la \wh{\ul{H}}_j \ra_{\ul{v}},
\end{equation}
for $\wh{\ul{H}}_i,\, \wh{\ul{H}}_j \in \mc{H}^{\sharp}$, then 
\begin{equation}
  \exists\, \ul{w}^{\sharp}
    \in \NN{\QCM{H^{\sharp}}{\psi^{\sharp}}}
  \tr{ s.t. }
  \tr{var}\, (\wh{\ul{H}}^{\sharp})_{\ul{\psi}^{\sharp}}
    = 0, 
\end{equation}
where $\wh{\ul{H}}^{\sharp} = (\mc{H}^{\sharp})_{\ul{w}^{\sharp}}$. 

Without having any knowledge of the number $T^{\sharp}$ of Hermitian operators to be used to generate $\wh{\ul{H}}^{\sharp}$, suppose our target space of Hamiltonians is $\tr{span}\, (\mc{H})$, where $\mc{H} = \{\wh{H}_i,\; i \in 0, \ldots, T - 1\}$ with $T < T^{\sharp}$. Then, we would have to first construct the QCM $\QCM{H}{\psi^{\sharp}}$ associated with $\mc{H}$ w.r.t. $\ul{\psi}^{\sharp}$ (see  Definition~\ref{def:QCM}), look for $\ul{w} \in \NN{\QCM{H}{\psi^{\sharp}}}$, and if such a $\ul{w}$ exists construct $\wh{\ul{H}} = (\mc{H})_{\ul{w}}$. Let us denote by $\{\ul{w}_n, \mu_n\},\; n \in 0, 1, \ldots, T - 1$, the eigen-pairs of $\QCM{H}{\psi^{\sharp}}$ ordered as $0 < \mu_0 \leq \mu_1 \leq \cdots \leq \mu_{T - 1}$. 

We have two cases to consider.


\subsection{Null Space of QCM Has Non-Zero Dimension}
\label{subsec:Dim1}


If 
\begin{align}
  \tr{dim}\, (\NN{\QCM{H}{\psi^{\sharp}}}) 
    \neq 0
    &\iff
       \ul{w}_0 
         \in \NN{\QCM{H}{\psi^{\sharp}}}
       \notag \\
    &\iff
       \QCM{H}{\psi^{\sharp}} \tr{ is p.s.d.}
       \notag \\
    &\iff
       \mu_0
         = 0,
  \label{eq:Dim1}
\end{align}
then $\wh{\ul{H}} = (\mc{H})_{\ul{w}} \in \tr{span}\, (\mc{H})$ has $\ul{\psi}^{\sharp}$ as an eigen-mode. 

If $\tr{dim}\, (\NN{\QCM{H}{\psi^{\sharp}}}) = 1$, there is only one such spin Hermitian operator in $\tr{span}\, (\mc{H})$ that has $\ul{\psi}^{\sharp}$ as an eigen-mode; if $\tr{dim}\, (\NN{\QCM{H}{\psi^{\sharp}}}) \geq 2$, there are multiple such operators. In this case, the question of which operator is most suitable arises. For practicality, we select the Hamiltonian that represents the KME in its lowest eigen-state to proceed. While algorithms can be employed to generate the sparsest basis for the null space \cite{Qu_2016}, determining the optimal Hamiltonian for UQ remains an open problem.

\begin{figure}[htpb]
    \centering
    \includegraphics[width = 0.95\columnwidth]{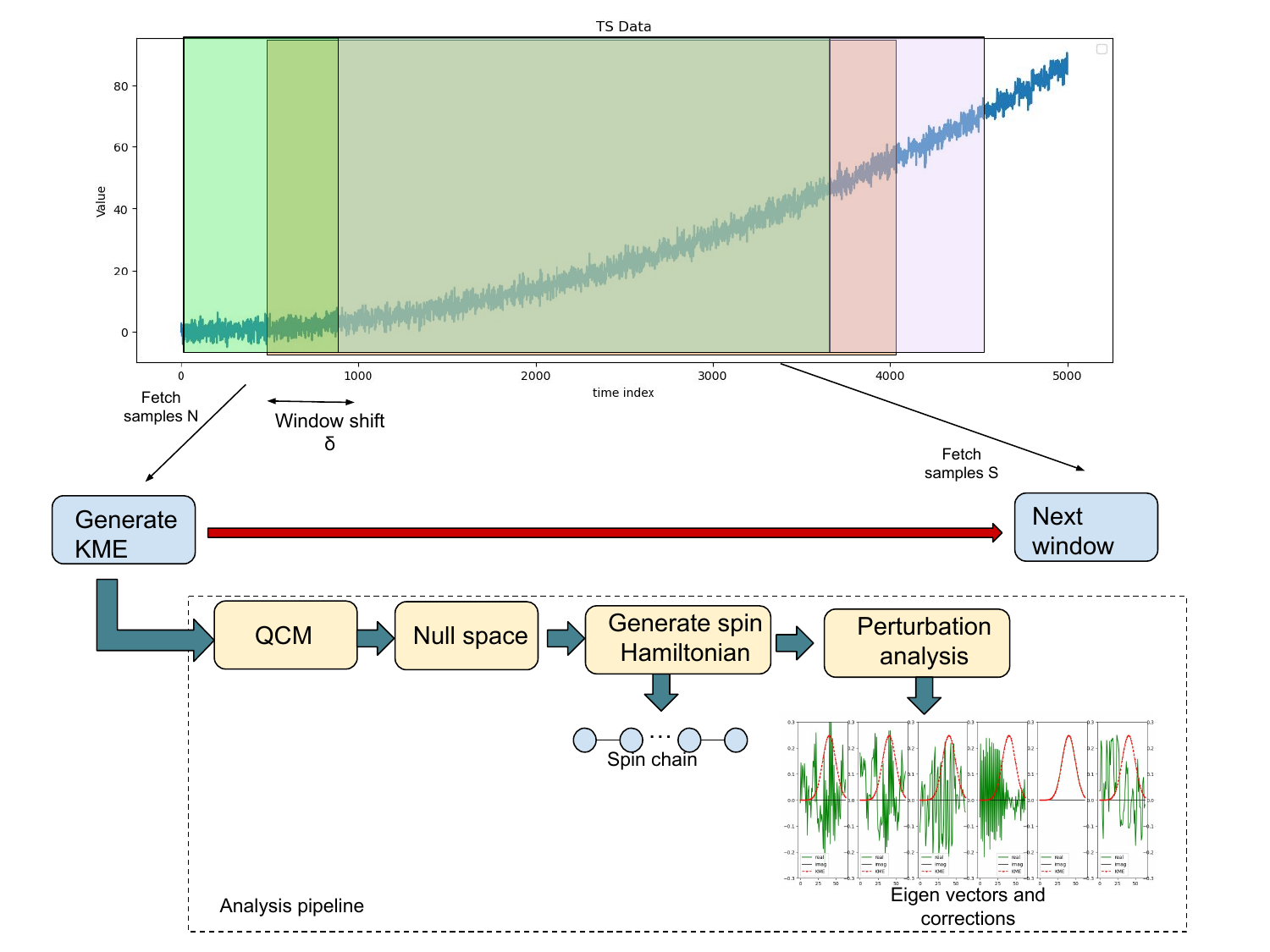}
    \caption{Pipeline of generating spin Hamiltonians from TS data and UQ.}
    \label{fig:UQ_pipeline}
\end{figure}

To deal with TS data, we take a moving window of fixed length and compute the KME for each such window. Figure~\ref{fig:UQ_pipeline} and Algorithm~\ref{alg:algorithm1} depict the pipeline and psudo-code for generating these spin Hamiltonians for TS data, respectively. Here, $\ell$ denotes the window index and $\delta$ denotes the stride length of the moving window.

\begin{figure}[ht]
\begin{algorithm}[H]
  \caption{Spin Chain Hamiltonian for TS Data}
  \label{alg:algorithm1}
  \begin{algorithmic}
    \STATE $\mc{H} = \{\wh{\ul{H}}_i,\; i \in 0, \ldots, T - 1\} \leftarrow$ Pairwise orthonormal target space of Hermitian operators. 
    \STATE $w[\ell] \leftarrow$ Window of size $N$ indexed as $\ell$.
    \STATE Fetch previous $N$ samples $\ul{y}[\ell](0:N - 1)$.
    \STATE \tb{Calculate KME $\ul{\psi}^{\sharp}[\ell]$ of $\ul{y}[n]$.}
    \STATE $\QCM{H}{\psi^{\sharp}} \leftarrow$ Generate QCM matrix w.r.t. $\ul{\psi}^{\sharp}[\ell]$ (Definition~\ref{def:QCM}).
    \STATE $w[\ell] \leftarrow$ Null space vector of $\QCM{H}{\psi^{\sharp}}$.
    \STATE $\wh{\ul{H}}[\ell] \leftarrow (\mc{H})_{\ul{w}[\ell]}$, spin chain Hamiltonian that has the KME $\ul{\psi}^{\sharp}[\ell]$ as an eigen-mode. 
    \STATE $\ul{\psi}_n[\ell],\; n \geq 0 \leftarrow$ Normalized eigen- vectors of $\wh{\ul{H}}[\ell]$.
    \STATE Shift $w[\ell]$ by $\delta$ number of samples.
    \STATE $\ell \leftarrow \ell + 1$
    \end{algorithmic}
\end{algorithm}
\end{figure}


\subsection{Null Space of QCM Has Zero Dimension}
\label{subsec:Dim0}


If 
\begin{align}
  \tr{dim}\, (\NN{\QCM{H}{\psi^{\sharp}}}) 
    = 0
    &\iff
       \NN{\QCM{H}{\psi^{\sharp}}} 
         = \{\ul{0}\} 
       \notag \\
    &\iff
       \QCM{H}{\psi^{\sharp}} \tr{ is p.d.}
       \notag \\
    &\iff
       \mu_0
         > 0,
  \label{eq:Dim0}
\end{align}
$\ul{\psi}^{\sharp}$ is not an eigen-mode of any Hermitian operator in $\tr{span}\, (\mc{H})$. In this case, for any given $\wh{\ul{H}} \in \tr{span}\, (\mc{H})$, $\tr{Var}\, (\wh{\ul{H}})_{\ul{\psi}^{\sharp}}$ quantifies how `close' $\ul{\psi}^{\sharp}$ is to being an eigen-mode of $\wh{\ul{H}}$. While we do not pursue this case in detail in this paper, we wish to state the following about how we can proceed. 

Consider a `perturbation' of the QCM $\QCM{H}{\psi^{\sharp}}$ given by 
\begin{equation}
  \ul{M}'_{\ul{\psi}^{\sharp}}
    = \QCM{H}{\psi^{\sharp}} - \delta\ul{M}
\end{equation}
s.t. $\tr{dim}\, (\NN{\ul{M}'_{\ul{\psi}^{\sharp}}}) = 1$ while keeping $\ul{M}'_{\ul{\psi}^{\sharp}}$ real-valued, symmetric, and p.s.d. For instance, one way to achieve this is to use
\begin{equation}
  \delta\ul{M}
    = \mu_0\, 
      \cdot
      \mid \ul{w}_0 \ra\, \la \ul{w}_0 \mid.     
\end{equation}
This would yield $\{\ul{w}_0, 0\} \ds\bigcup_{n = 1}^{T - 1} \{\ul{w}_n, \mu_n\}$ as the eigen-pairs of $\ul{M}'_{\ul{\psi}^{\sharp}}$. In particular, $\ul{w}_0 \in \NN{\ul{M}'_{\ul{\psi}^{\sharp}}}$ and since $\tr{dim}\, (\NN{\ul{M}'_{\ul{\psi}^{\sharp}}}) = 1$, we must have $\NN{\ul{M}'_{\ul{\psi}^{\sharp}}} = \tr{span}\, \{\ul{w}_0\}$. But the difficulty here lies in that there is no guarantee that the perturbed $\ul{M}'_{\ul{\psi}^{\sharp}}$ is  associated with a set $\mc{H}'$ of mutually orthogonal Hermitian operators of the type in Section~\ref{sec:TargetSpace}. In other words, there is no gauarantee that $\ul{M}'_{\ul{\psi}^{\sharp}}$ is a `valid' QCM.

If such a set $\mc{H}'$ exists, then one can employ $\ul{w}_0$ as the linear combination weights to generate the Hermitian operator being sought as $\wh{\ul{H}}' = (\mc{H}')_{\ul{w}_0}$ and $\ul{\psi}^{\sharp}$ is guaranteed to be an eigen-mode of $\wh{\ul{H}}'$. Unfortunately, the more likely scenario is that such a set $\mc{H}'$ does not exist. Then, one can still get $\wh{\ul{H}}'$ as $\wh{\ul{H}}' = (\mc{H}')_{\ul{w}_0}$ but it will not have $\ul{\psi}^{\sharp}$ as an eigen-mode. One may then attempt to quantify and bound the error between $\ul{\psi}^{\sharp}$ and the eigen-mode associated with the smallest eigen-energy of $\wh{\ul{H}}'$ \cite{Qi2019Quantum}. 


\section{Uncertainty Quantification (UQ)}


In the work presented in \cite{singh2020time, singh2021toward}, which is formulated based on the QIPF in \cite{Principe2010_Book}, the eigen-modes are pre-determined in that they are the set of Hermite polynomials which constitutes an ONB for the QHO. Therefore, it is the `projection' of the KME $\ul{\psi^{\sharp}}$ onto this set of Hermite polynomials that constitutes the linchpin of the methodology in \cite{singh2020time, singh2021toward}. Using $\wh{\ul{\psi}}_n(x),\; n \geq 0$, to denote these  projections, \cite{singh2020time, singh2021toward} employ
\begin{multline}
  V_n(x) 
    = \wh{E}_n + \frac{\sigma^2}{2}\, \frac{\nabla_n^2 |\wh{\ul{\psi}}_n(x)|}{|\wh{\ul{\psi}}_n(x)|}, \\
  \tr{where }
  \wh{E}_n 
    = - \min_x
        \frac{\sigma^2}{2}\, \frac{\nabla_n^2 |\wh{\ul{\psi}}_n(x)|}{|\wh{\ul{\psi}}_n(x)|},
  \label{eq:rishab_e}
\end{multline}
for different tasks (including change point detection, UQ, etc.). It captures the rate of change of the projections $\wh{\ul{\psi}}_n(x)$ across different modes, with rapid changes indicating higher uncertainty. The eigen- energy term $\wh{E}_n$ is used as a `bias' to ensure that $\min\, V_n(x) = 0$. This approach then offers a localized measure of uncertainty which adapts to the specific structure of the data. 

(b)~In our work, each spin chain associated with the given KME $\ul{\psi}^{\sharp}$ generates its own `customized' spin Hamiltonian $\wh{\ul{H}} \overset{\Delta}{=} (\mc{H})_{\ul{w}_0}$, where $\ul{w}_0 \in \NN{\QCM{H}{\psi^{\sharp}}}$ (see Section~\ref{subsec:Dim1}). Perturbation theory is applied to the set of eigen-modes of this Hamiltonian $\wh{\ul{H}}$. 

The metric we employ to quantify uncertainty is motivated by the same measure \eqref{eq:rishab_e}, except that the projection terms $\wh{\ul{\psi}}_n(x)$ are replaced by the first-order corrections generated by perturbation theory. To be more specific, corresponding to \eqref{eq:rishab_e}, we employ
\begin{multline}
  V_n^{(1)}(x)
    = E_n^{(1)} + \frac{\sigma^2}{2}\, \frac{\nabla_n^2 |\ul{\psi}_n^{(1)}(x)|}{|\ul{\psi}_n^{(1)}(x)|}, \\
  \tr{where }
  E_n^{(1)} 
    = -\min_x
       \frac{\sigma^2}{2}\, \frac{\nabla_n^2 |\ul{\psi}_n^{(1)}(x)|}{\ul{\psi}_n^{(1)}(x)|},
  \label{eq:UQ_equation}
\end{multline}
where $\{\ul{\psi}_n,\, E_n\},\; n \geq 0$, denotes the eigen-pairs of the spin Hamiltonian $\wh{\ul{H}}$ ordered as $\lambda_0 \leq \lambda_1 \leq \cdots$, and $(\cd)^{(1)}_n$ denotes the first-order correction terms associated with the $n$-th eigen-pair when a small perturbation is applied onto the Hamiltonian \cite{CohenTannoudji1977}. 

\begin{figure}[htbp]
  \centering
  \label{fig:sinewave}
  \includegraphics[width = 0.80\columnwidth]{%
    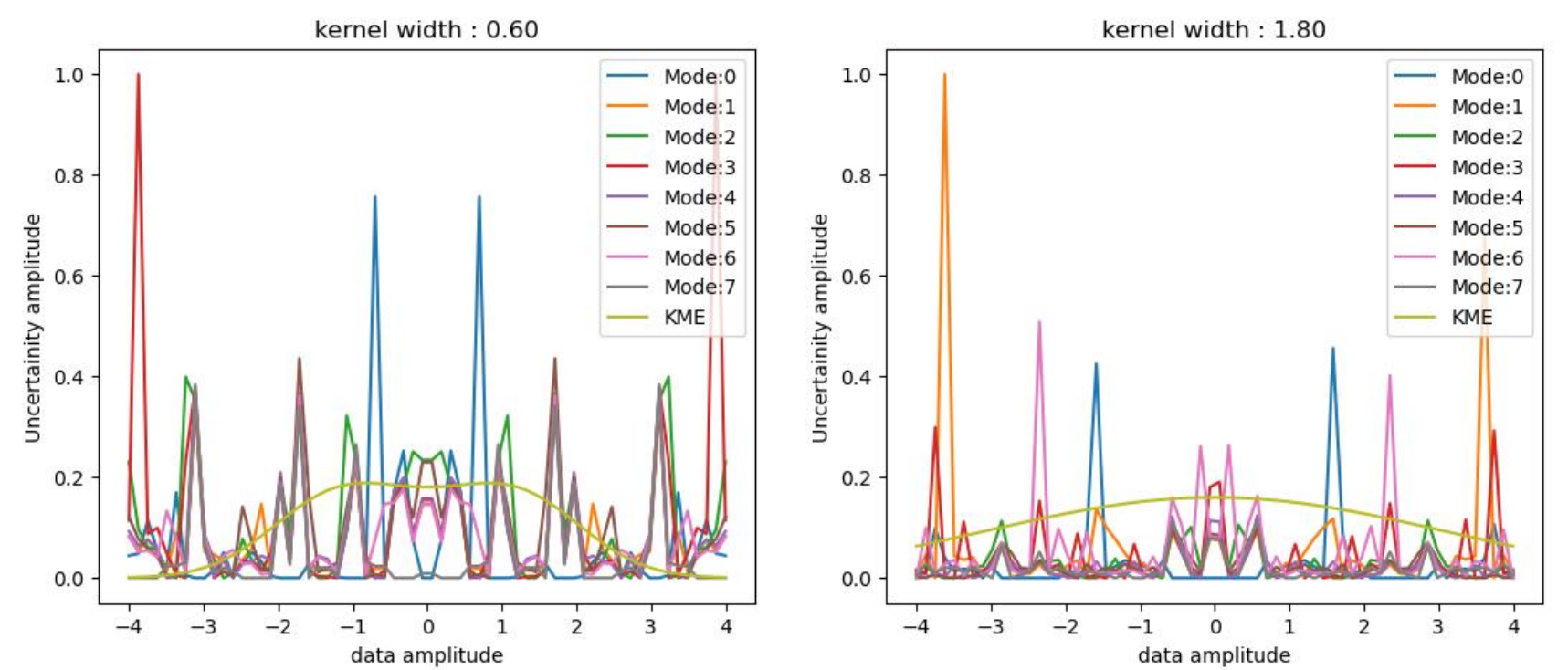}
  \caption{Uncertainity quantification of the sine wave in the space of data for two different kernel widths.}
\end{figure}

Next, we demonstrate how our proposed framework characterizes uncertainty of a simple TS signal. For this, we take the same simple illustrative example that appears in \cite{singh2021toward} where the data vector is a 3000-sample $50\, \tr{(Hz)}$ sinusoid sampled at a frequency of $6000\, \tr{(samples/s)}$ normalized
to zero mean and unit standard deviation. Utilizing all 3000 samples as `centers', a KME within the domain $x = (-4,\, 4)$ containing 64 samples was generated using \eqref{eq:KME}. With this KME, the spin chain Hamiltonian was determined via the QCM in Definition~\ref{def:QCM} and \eqref{eq:LinearCombination}. The smallest 8 eigen-modes (meaning those associated with the smallest 8 eigen-energies) of the Hamiltonian was used for UQ utilizing \eqref{eq:UQ_equation}. We repeated the experiment for two different kernel widths and Figure~\ref{fig:sinewave} shows the results obtained. All plots were normalized for easier visualization. One can observe that the uncertainty peaks of various eigen-modes align with regions of sharp data PDF transitions as well as the tails of the data PDF. This is in agreement with what is observed in \cite{singh2021toward}.

\section{Experiments}
\label{sec:Experiments}

In this section we provide details on the experiments that we conducted to assess the performance of our proposed methodology. These involve change point detection (both mean jump and variance jump) and TS clustering. All simulations were executed using Python~3.8. For all experiments the number of orthonormal target space was maintained at $T=363$ and spin chain length was set to $L=6$.


\subsection{Change Point Detection}


\subsubsection{Datasets}


For change point detection, we followed the procedure in \cite{takeuchi2006unifying} to generate artificial datasets where change points were manually inserted at specific intervals to induce statistical shifts. We generated two  datasets, one where change points simulate shifts in mean and the other which simulate shifts in variance. 

\begin{figure}[ht]
  \centering
  \subfloat[%
    \tr{Mean jumps data.}]{%
    \label{fig:Mean jump data}
    \includegraphics[width = 0.35\columnwidth]{%
      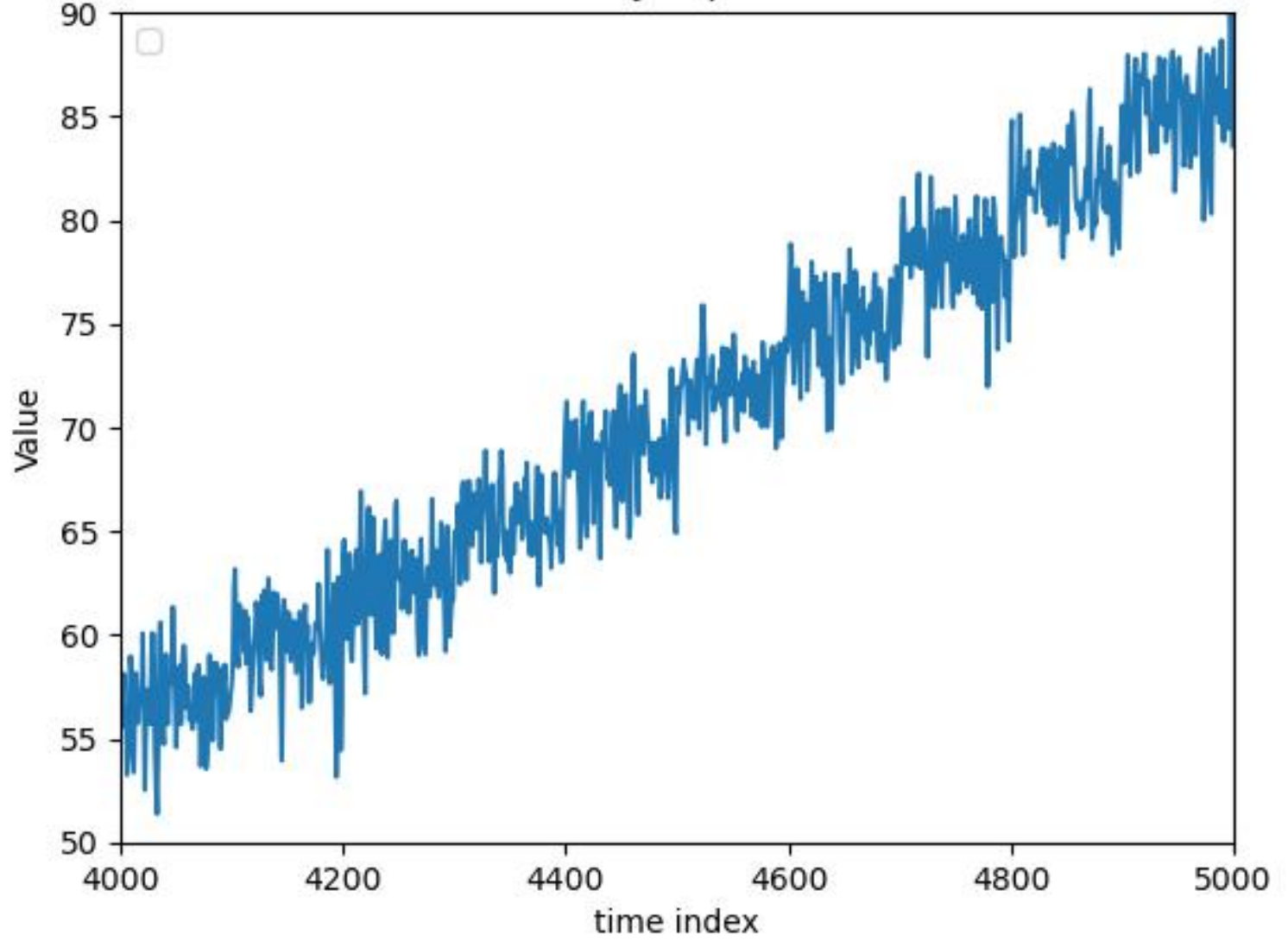}} 
  \subfloat[%
    \tr{Variance jumps data.}]{%
    \label{fig:Variance Jump data}
    \includegraphics[width = 0.35\columnwidth]{%
      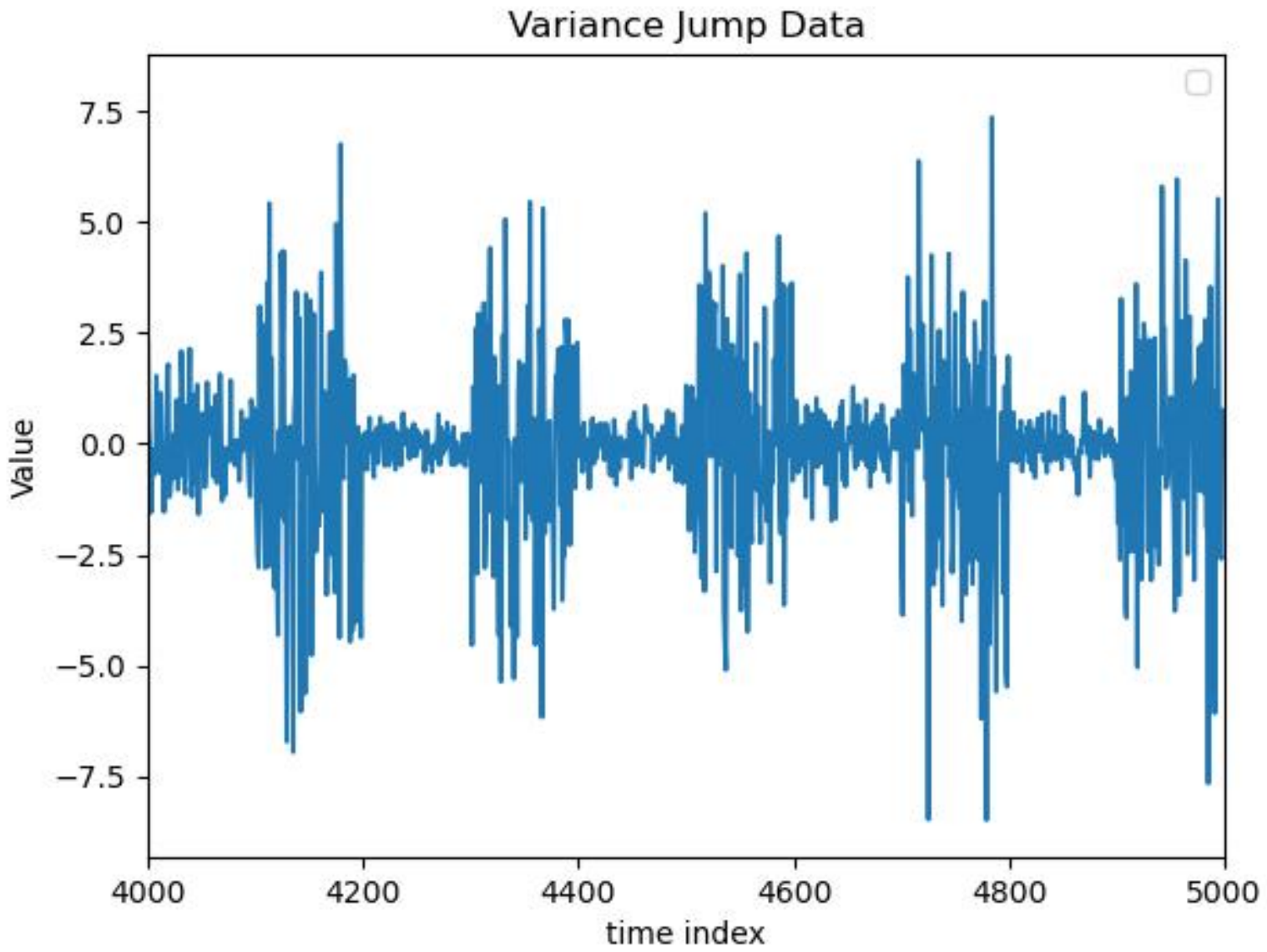}} \\
  \subfloat[%
    \tr{Mean jumps detection ROC.}]{%
    \label{fig:ROC Mean jump data}
    \includegraphics[width = 0.35\columnwidth]{%
      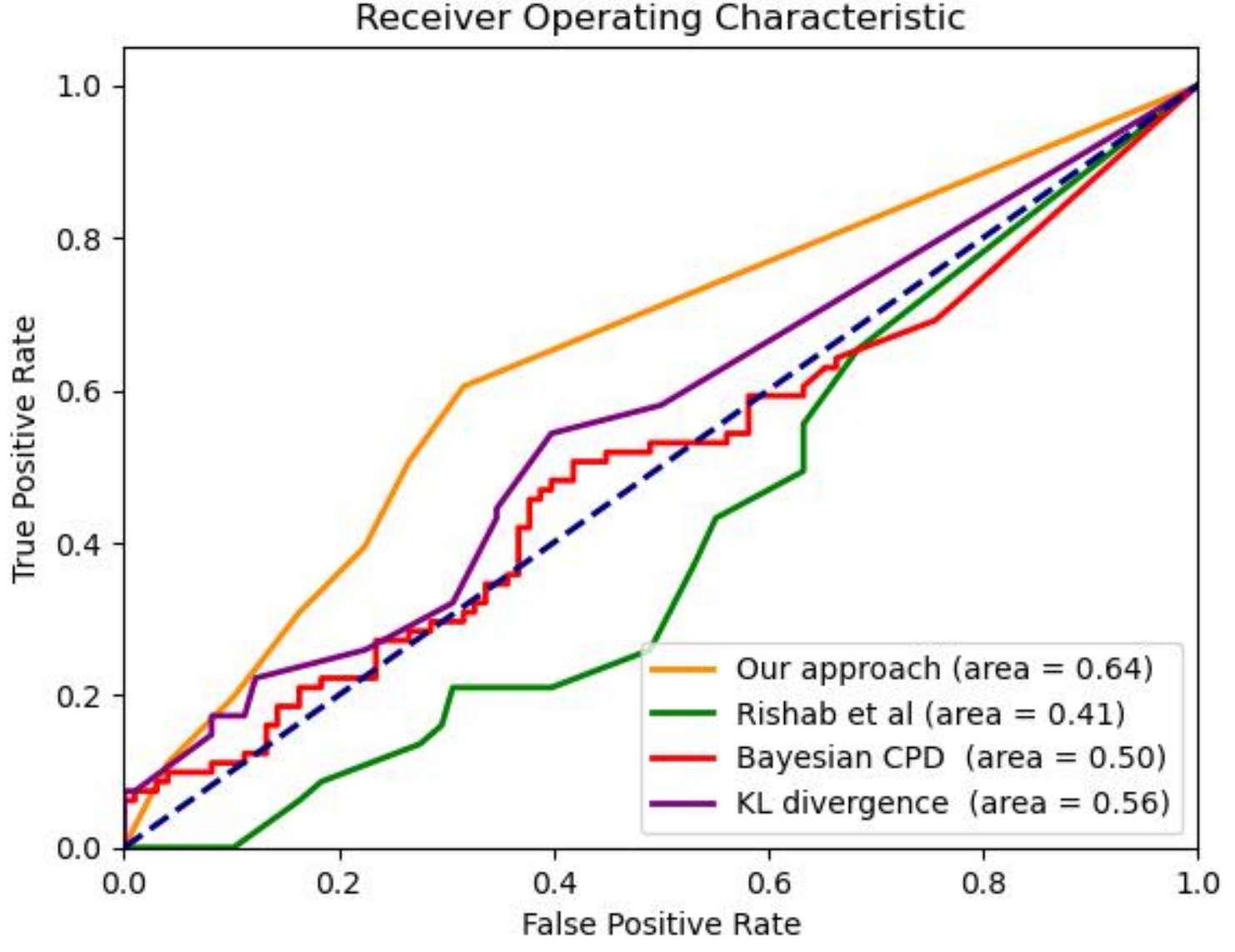}}
  \subfloat[%
    \tr{Variance jumps detection ROC.}]{%
    \label{fig:ROC Variance jump data}
    \includegraphics[width = 0.35\columnwidth]{%
      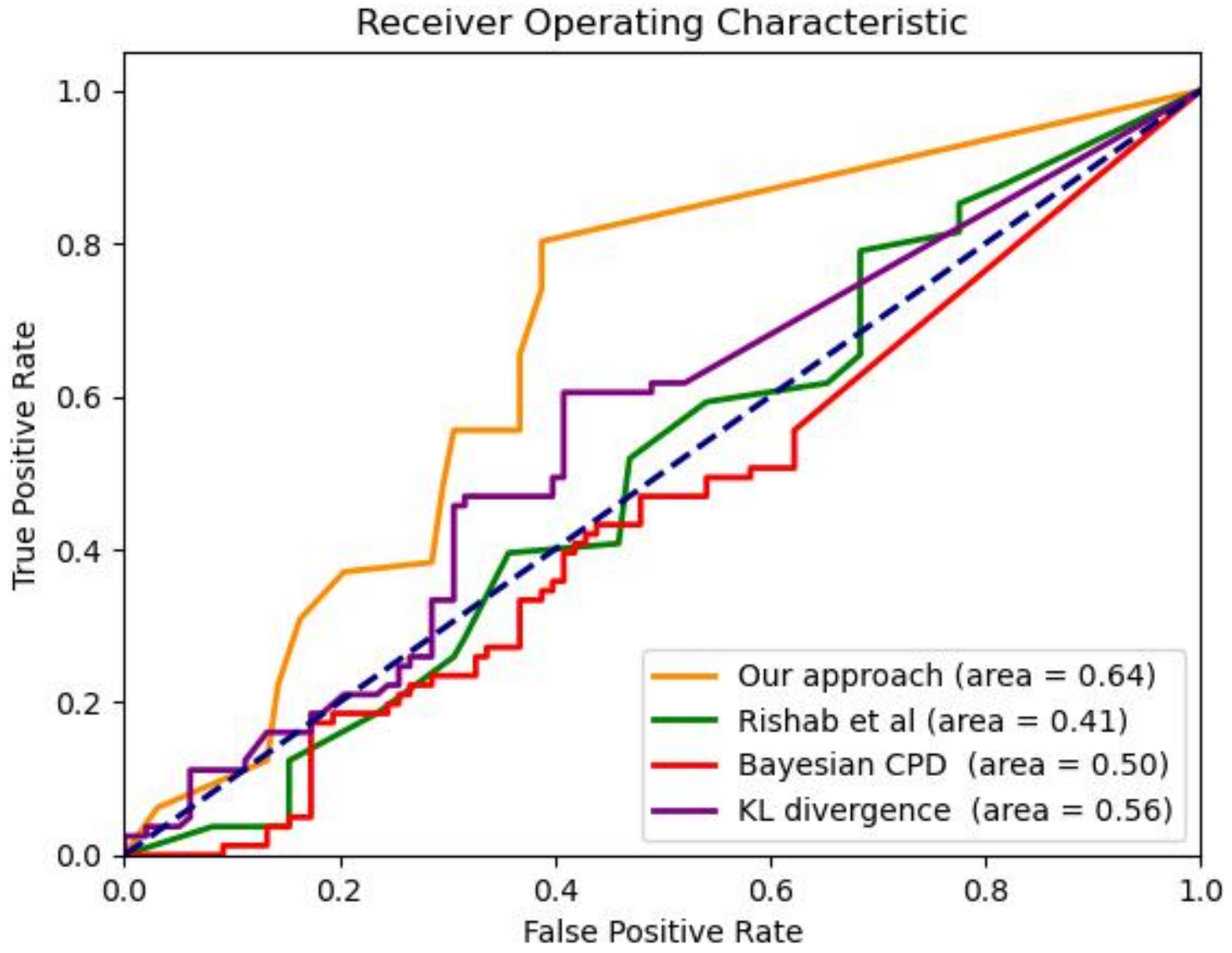}}  
  \caption{Top row: last 1000 samples of change point datasets. Bottom row: ROC curves for different methods measured over the same range of samples for both datasets.}
  \label{fig:change_points}
\end{figure}

\textbf{Mean Jumps Dataset.} 
To simulate a TS with mean jumps at regular intervals, we synthesized 5000 samples using the auto-regressive model
\begin{equation}
  y(t) 
    = 0.6\, y(t - 1) - 0.5\, y(t - 2) + \epsilon_t,
\end{equation}
with initial conditions $y(1) = y(2) = 0$. Here, $\epsilon_t$ represents Gaussian noise with mean $\mu$ and standard deviation $\sigma$ kept at $1.5$. We introduce a change point at every 100 time samples by adjusting the noise mean $\mu$ at time $t$ as
\begin{equation}
  \mu_N 
    = \begin{cases} 
        0,
          & \tr{for $N = 1$}; \\
        \mu_{N - 1} + N/16,
          & \tr{for $N = 2,\, \ldots,\, 49$}.
      \end{cases}
\end{equation}
Here, $N$ is a natural number such that $100\, (N - 1) + 1 \leq t \leq 100\, N$. The purpose of synthesizing such a dataset is to create drifts in the data without making the change points immediately apparent to the human eye, thereby posing a challenging detection task for the algorithm. See Figure~\ref{fig:Mean jump data}. 

\textbf{Variance Jumps Dataset.} 
Similar to the mean jumps dataset, we introduce a change point at every 100 time steps in the noise standard deviation $\sigma$, while maintaining the noise mean $\mu$ at $0$, via the model
\begin{equation}
  \sigma_N 
    = \begin{cases} 
        \mr{random}\, (0, 1) 
          & \tr{for $N$ is odd}; \\
        \ln\, (e + N/4),
          & \tr{for $N$ is even}.
      \end{cases}
\end{equation}
Here, when $N$ is odd, variance is randomly uniformly chosen between $0$ and $1.0$; when $N$ is even, it is fixed at $\ln\, (e + N/4)$. See Figure~\ref{fig:Variance Jump data}.


\subsubsection{Procedure}


We computed the KME as in \eqref{eq:KME} with $\sigma$ set to five times Silverman's rule of thumb (SROT) \cite{silverman2018density}. Algorithm~\ref{alg:algorithm1} was used to compute the spin chain Hamiltonian and its eigen-modes. 

To compare the performance of our framework, we used the online Bayesian change point detector \cite{adams2007bayesian}, the Kullback-Leibler divergence measure \cite{kullback1951information} and the work in \cite{singh2020time} which employs the QIPF \cite{Principe2010_Book}. These algorithms were selected due to their non-parametric and unsupervised nature. 

\begin{figure}[htpb]
  \centering
   \includegraphics[width = 0.65\columnwidth]{%
      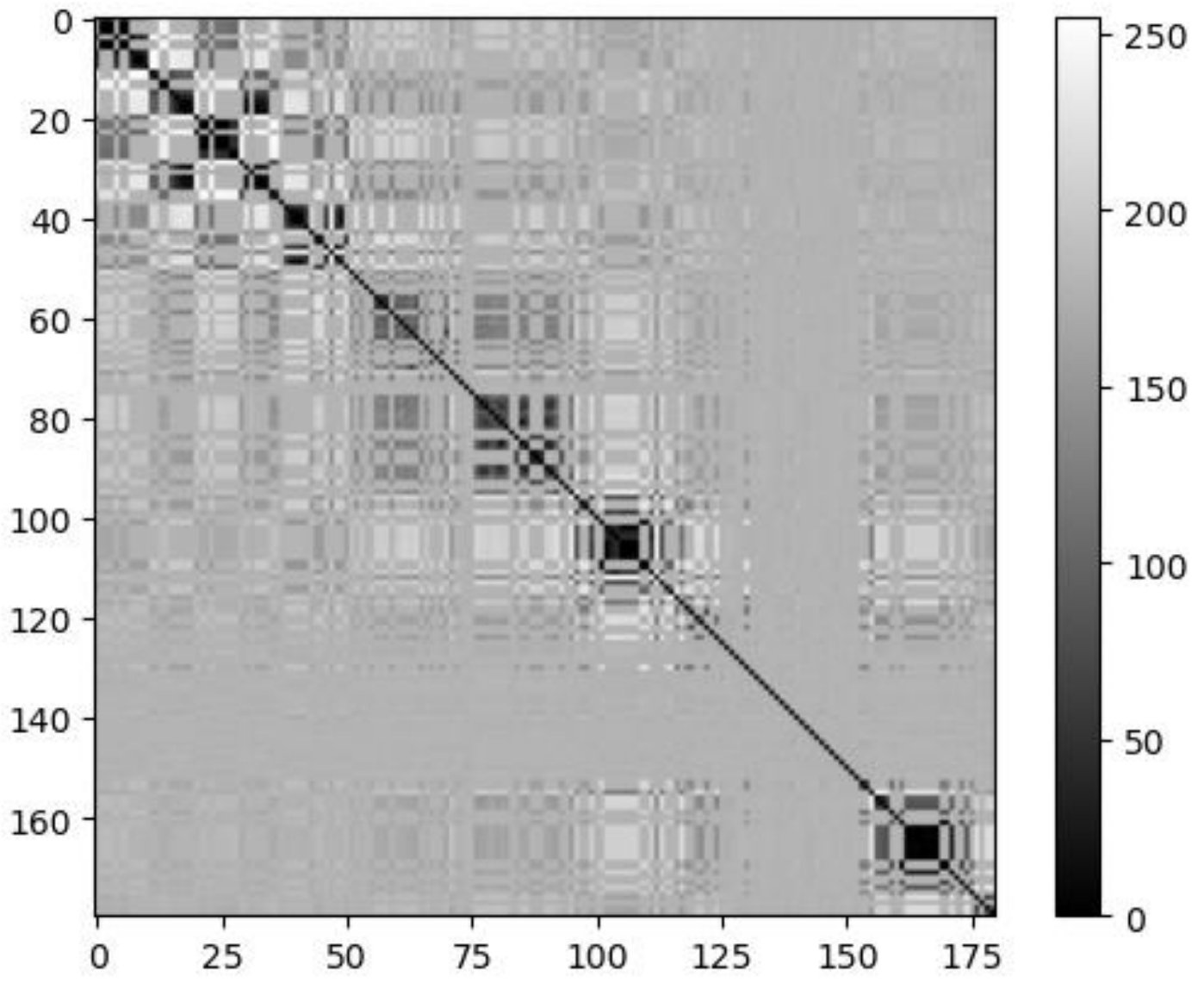}
  \caption{Mean jumps detection experiment: mean Euclidean distance of the eigen-modes of spin chain Hamiltonians between pairs of windows.}
  \label{fig:euclidian_distance}
\end{figure}

Our hypothesis is that changes in the TS characteristics will manifest as changes in the spin Hamiltonian's eigen-modes. Specifically, these changes should be reflected in the increased Euclidean distance between the eigen-modes of consecutive windows. For the mean jumps dataset, Figure~\ref{fig:euclidian_distance} shows the mean Euclidean distance of the eigen-modes between pairs of windows. To detect the changes in the Euclidean distance between windows, we employed a local maxima detection scheme from the SciPy standard library \cite{2020SciPy-NMeth}. 

The ROC curves shown in Figures~\ref{fig:ROC Mean jump data} and \ref{fig:ROC Variance jump data} which show the true positive rates against the false positive rates for the four methods over a range of thresholds from 4000 to 5000 samples of the datasets demonstrate that the spin chain Hamiltonian framework outperforms the other methods in detecting mean jump and variance jump changes.


\subsubsection{UQ}

To quantify the uncertainty regarding the detected locations of change points, we employ the following method: take $w[\ell]$, the TS window indexed as $\ell$; use Algorithm~\ref{alg:algorithm1} to compute the KME $\ul{\psi}^{\sharp}[\ell]$ and the spin Hamiltonian $\wh{\ul{H}}[\ell]$ associated with it; apply perturbation theory to the eigen-modes of $\wh{\ul{H}}[\ell]$ and compute their first-order corrections $\{\ul{\psi}_n^{(1)}(x)[\ell],\, E_n^{(1)}[\ell]\},\; n \geq 0$; and then use the expectation 
\begin{equation}
  \tr{UQ}[\ell]
    = \lla
        V_n^{(1)}(x)[\ell]
      \rra_{\ul{x}}
    \overset{\Delta}{=}
      \la \ul{x} \mid V_n^{(1)}(x)[\ell] \mid \ul{x} \ra.
  \label{eq:UQ_equation2}
\end{equation}
Note that $\tr{UQ}[\ell]$ is a non-negative real number which captures the uncertainty associated with window $w[\ell]$. This process is then repeated over successive windows. 

Such an approach provides a robust metric for UQ, which is crucial for quantifying the confidence one may place on the locations of detected change points. The  measure in \eqref{eq:UQ_equation2} helps identify false positives and negatives more effectively, providing better insight into the performance and robustness of the change point detection method. 

\begin{figure}[ht]
  \centering
  \subfloat[%
    \tr{Mean jumps data.}]{%
    \label{fig:UQ_Mean}
    \includegraphics[width = 0.80\columnwidth]{%
      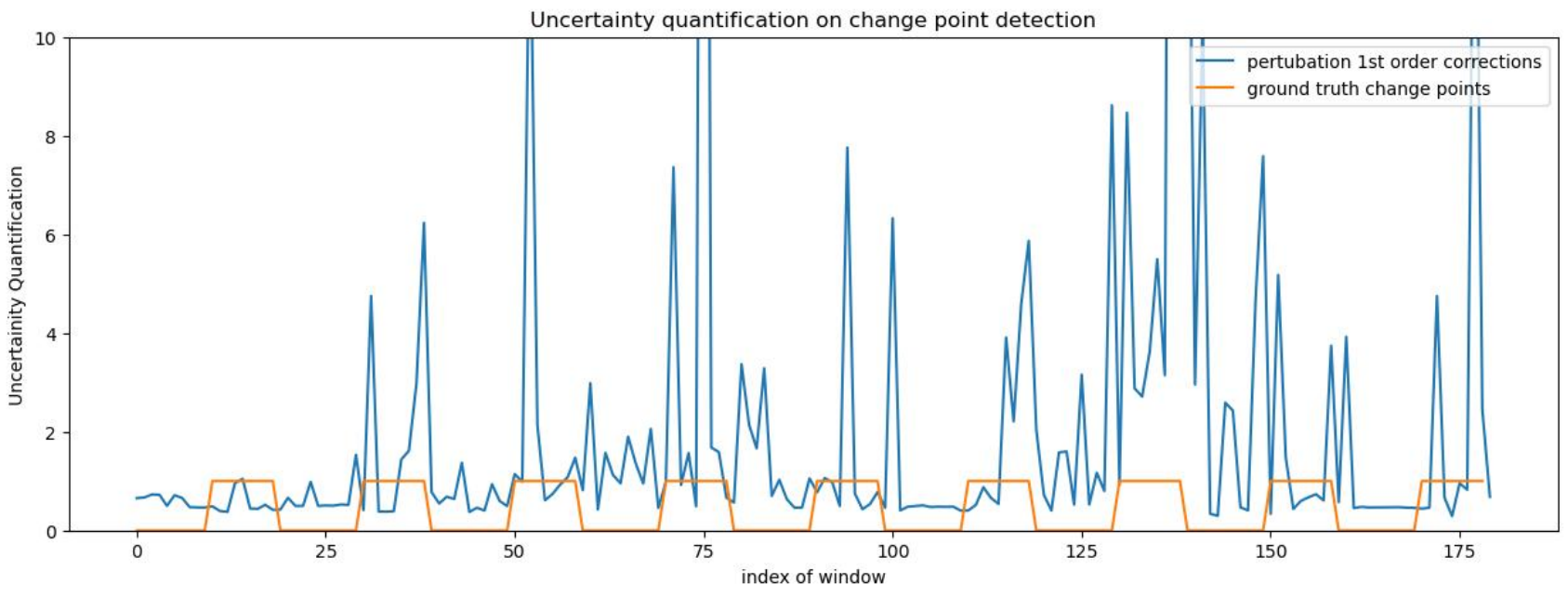}} \\
  \subfloat[%
    \tr{Variance jumps data.}]{%
    \label{fig:UQ_STD}
    \includegraphics[width = 0.80\columnwidth]{%
      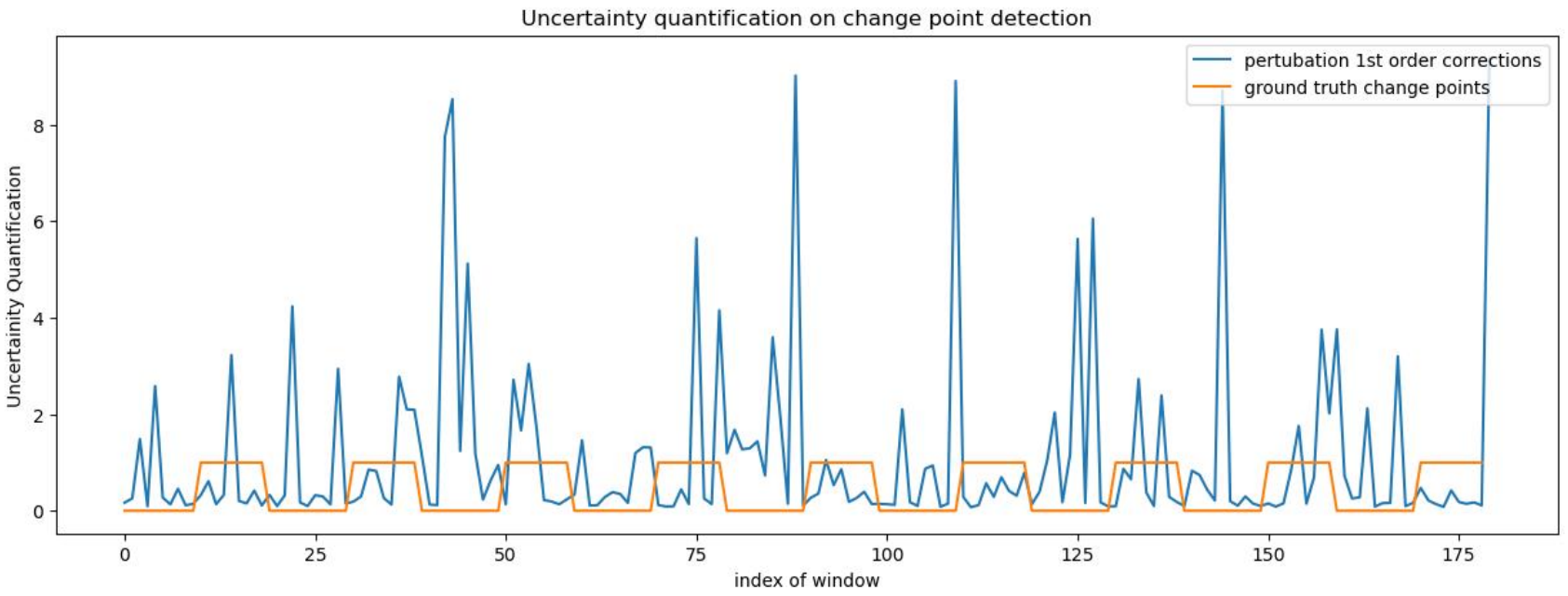}}
  \caption{UQ regarding the detected locations of change points. The ground truth locations of mean/variance jumps are in orange.}
  \label{fig:UQ}
\end{figure}

Figures~\ref{fig:UQ_Mean} and \ref{fig:UQ_STD} show the results we get from our mean jumps and variance jumps datasets. The ground truth locations of mean/variance jumps are in orange. One notes that in regions where the mean/variance jumps occur, the uncertainty exhibited is higher. 

\subsection{TS Clustering}


\subsubsection{Dataset}


For purposes of TS clustering, we assessed our framework using a subset of the VidTIMIT dataset, which includes voice recordings from 43 different speakers \cite{sanderson2009multi}. The dataset poses a significant challenge due to the presence of head movements during recording. 


\subsubsection{Procedure}


Following a method similar to \cite{singh2020time}, we randomly select 5 speakers, each providing 5 unique voice recordings. This results in a total of 25 TS datasets. Our goal was to cluster these datasets into their corresponding speaker classes in an unsupervised manner. To prepare the data, we downsampled each signal by a factor of 20 and used only the middle 2000 samples for our experiments.

This evaluation aims to determine which feature set maximizes the distance between different sentence classes while minimizing intra-class distances.
We create a histogram vector for each TS signal, capturing the frequency of dominance across the KME domain. This histogram vector serves as the feature vector. 

To compare the performance of our framework, we use a method that employs the discrete wavelet transform (DWT) \cite{edwards1991discrete} and the work in \cite{singh2020time}. DWT was chosen as one of the baselines because it is a well-established method for unsupervised feature extraction in TS signal processing. It effectively utilizes both spatial and frequency components to extract features. Conversely, the QIPF framework extracts features as eigen-states of a quantum oscillator Hamiltonian, offering a unique approach to feature extraction.

Experiments are conducted using z-normalized data, with the kernel width set to 5 times SROT for the spin chain Hamiltonian and 40 times SROT for the QIPF framework. For DWT, we apply maximum level decomposition using the Daubechies-2 wavelet\cite{daubechies1992ten} to extract coefficients.

\begin{figure}[ht]
  \centering
    \subfloat{%
    \label{fig:euclidian_distance_methods}
    \includegraphics[width = 0.98\columnwidth]{%
      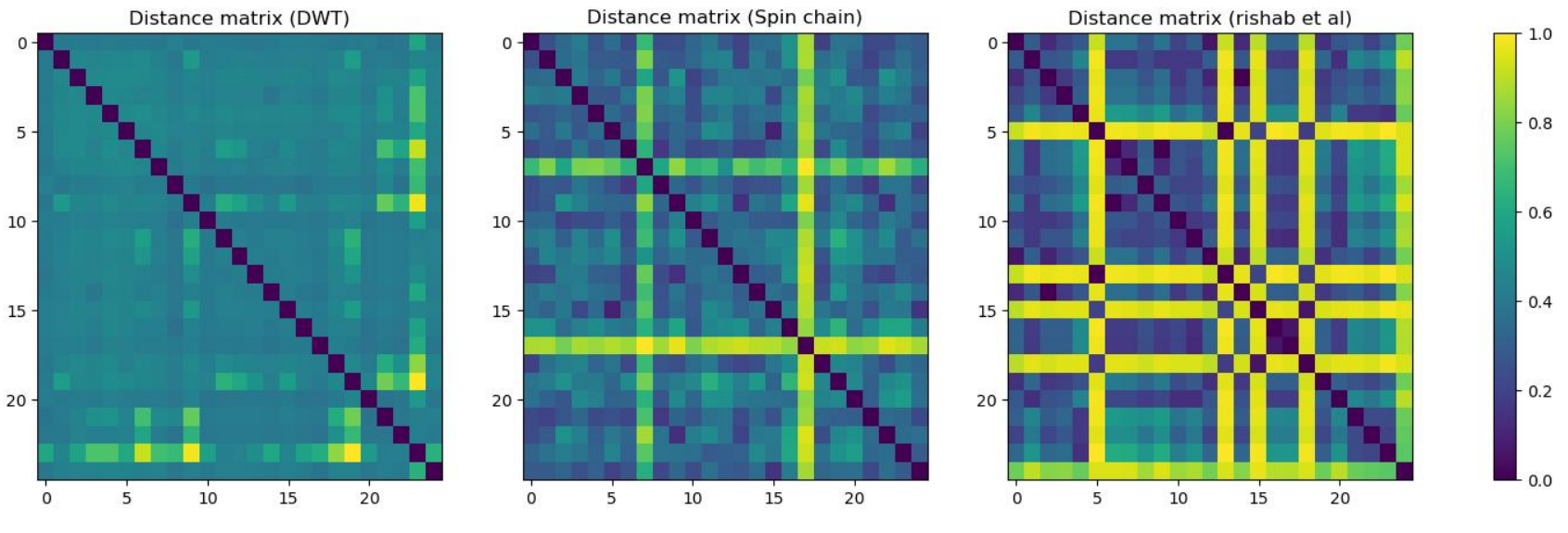}} \\
  \subfloat{%
    \label{fig:dendogram}
   \includegraphics[width = 0.98\columnwidth]{%
      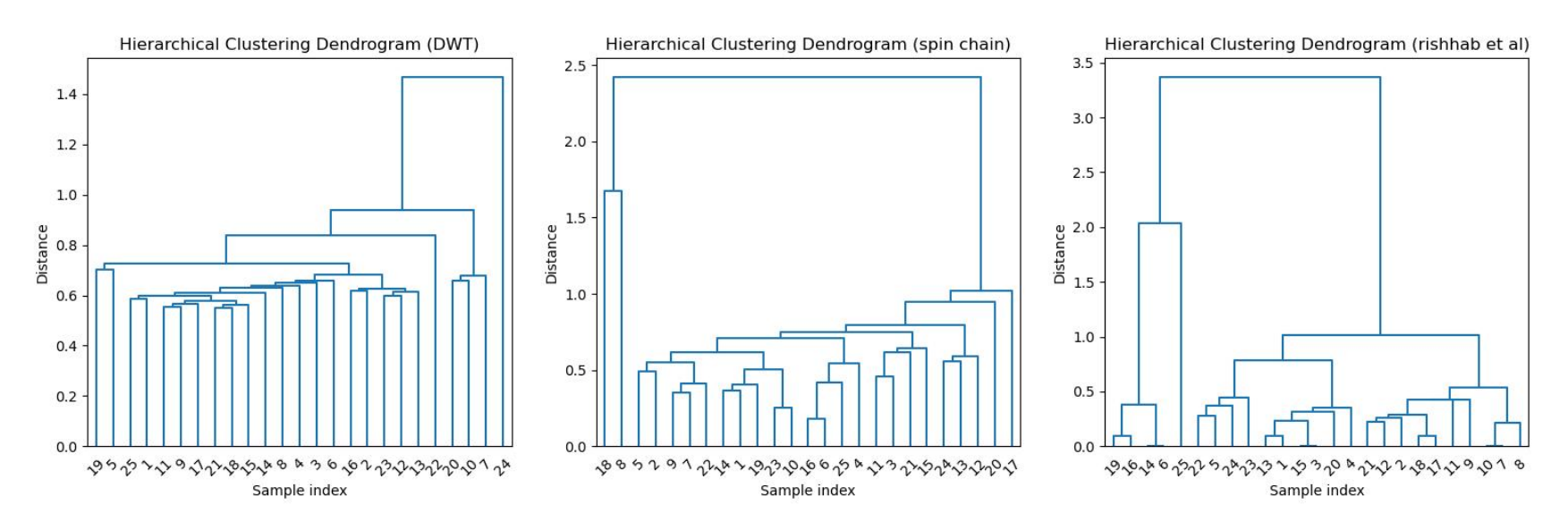}} \\
  \caption{Heatmaps and corresponding dendrograms depict pairwise feature vector distances for each of the 25 voice files. These visualizations compare the DWT (left), the spin chain framework (center), and the QIPF framework (right).}
  \label{fig:dendogram_and_heatmap}
\end{figure}

To evaluate the quality of the extracted features, we compute pairwise Euclidean distances between the features of different TS signals for each method. This helps to determine how closely the features of each signal matched its true class. The resulting heatmaps, which display the relative Euclidean distances between the different signals, are shown in Fig. \ref{fig:euclidian_distance_methods}. Additionally, hierarchical clustering is performed on the pairwise distances represented by the heatmaps to assess the clustering quality achieved by DWT, the spin chain framework, and QIPF.

\subsubsection{Clustering Performance}


To evaluate the effectiveness of hierarchical clustering, we employed three different metrics, each providing insight on clustering quality.
\begin{itemize}
  \item \emph{Cophenetic correlation coefficient~(CCC)} \cite{sokal1962comparison} measures how faithfully a dendrogram preserves the pairwise distances between the original data points. A higher CCC indicates that the hierarchical clustering structure accurately reflects the dataset's intrinsic similarities and dissimilarities. 
  \item \emph{Silhouette score (SS)} \cite{rousseeuw1987silhouettes} evaluates how well-separated the clusters are by comparing the average intra-cluster distance to the average nearest-cluster distance for each point. A higher SS indicates well-defined  distinct clusters. 
  \item \emph{Adjusted Rand index (ARI)} \cite{hubert1985comparing} assesses the similarity between the clusters produced by the algorithm and a ground truth classification, adjusting for chance. Higher ARI values signifying better alignment with the true clusters. 
\end{itemize}
the class of sentence provided by the dataset was taken as the ground truth classification. To determine which feature set yields the best clustering results, we compare the values of these metrics across the different methods. As Table~\ref{tab:1} domnstrates, the spin chain Hamiltonian appears to offer better TS clustering when compared with the DWT and QIPF frameworks. 

\begin{table}[ht]
  \caption{Clustering Performance}
  \centering
  \begin{tabular}{c ccc}
    \hline
    \tb{Evaluation Metric} 
      & \tb{DWT \cite{heil1989continuous}}
      & \tb{QIPF \cite{singh2020time}}
      & \makecell{\tb{Proposed} \\ \tb{Method}} \\ 
    \hline
    CCC 
      & 0.92 & \tb{0.97} & 0.95 \\
    \hline
    SS
      & 0.014 & 0.12 & \tb{0.13}\\
    \hline
    ARI
      & 0.047 & -0.043 & \tb{0.086}\\
    \hline
  \end{tabular}
  \label{tab:1}
\end{table}


\subsubsection{UQ}

\begin{figure}[ht]
  \centering
  \includegraphics[width = 0.80\columnwidth]{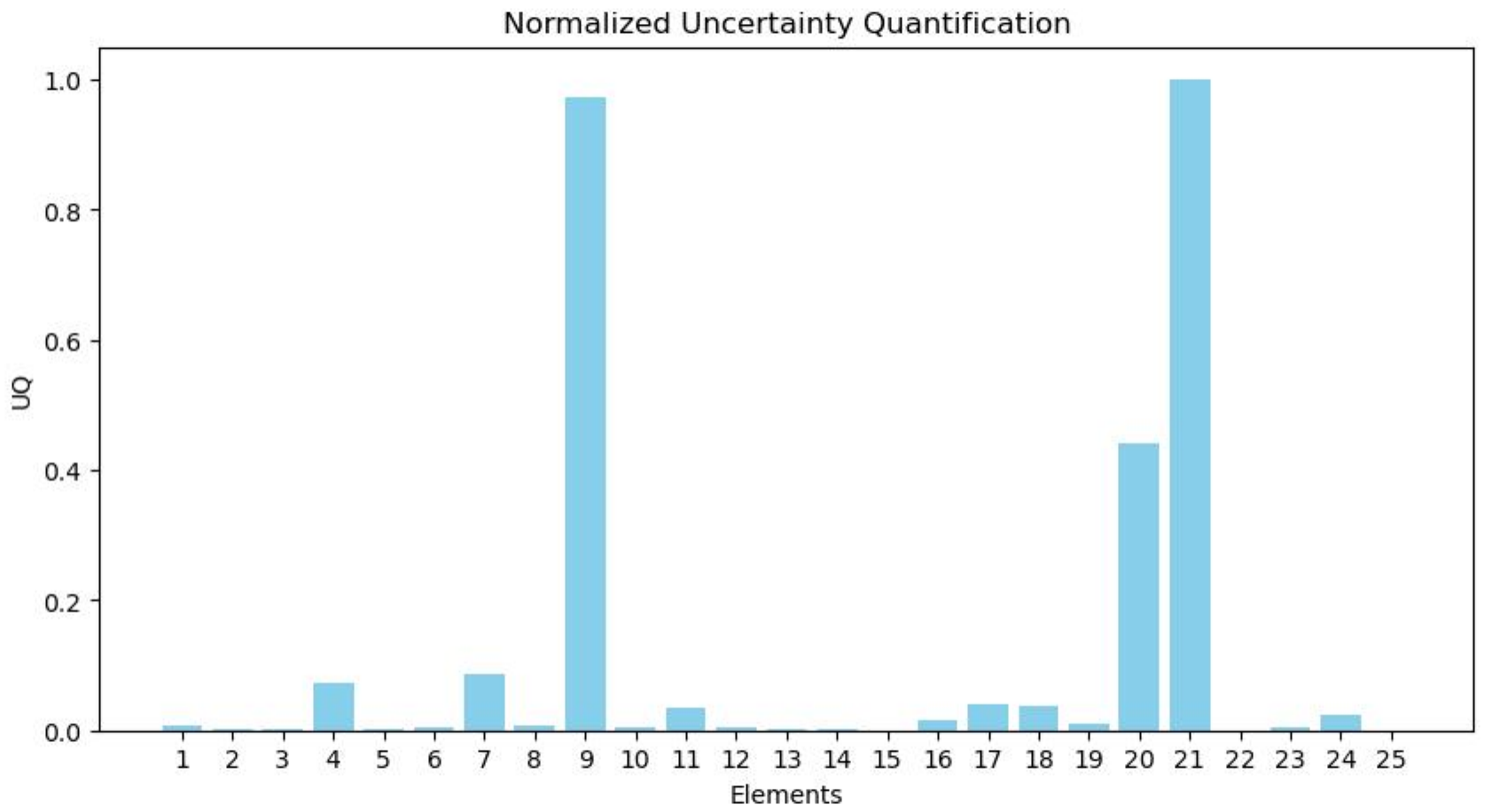}
  \caption{Sensitivity analysis of clustering}
  \label{fig:UQ_cluster}
\end{figure}

The uncertainty of each element subjected to the clustering was also calculated utilizing \eqref{eq:UQ_equation2}. 
Cluster member that are more uncertain in the sense of their membership can be identified from the normalized mean UQ given in Figure~\ref{fig:UQ_cluster}. Elements with higher uncertainty indicate less confidence about their cluster assignment. These elements may represent borderline cases or outliers that are difficult to categorize confidently.

\section{Conclusion}


We have proposed a quantum tensor network-based model for time series data with the objective of leading to more interpretable machine learning models, where the confidence in the decisions can be quantified. The framework is designed to extract sensitive uncertainty features, which are particularly useful for change point detection and TS clustering. We validate the effectiveness of our approach using both synthetic and real-world datasets. Our results show that the spin chain Hamiltonian framework significantly outperforms current state-of-the-art non-parametric methods. Additionally we provide a method to quantify the uncertainty associated with the data via perturbation analysis. 
While this paper presents a promising quantum tensor network-based model for time series data, several important questions remain unanswered. For instance, if the null space dimension of the Quantum Covariance Matrix (QCM) exceeds 1, multiple spin Hamiltonians could arise, each possessing the Kernel Mean Embedding (KME) as an eigen-mode. The challenge lies in determining which Hamiltonian to choose. Additionally, in scenarios where the null space dimension of the QCM is zero, we have referenced the method proposed in \cite{Qi2019Quantum} as one potential solution (see Section~\ref{subsec:Dim0}). An alternative approach, which we have yet to fully explore, involves utilizing the orthogonality principle as it pertains to optimal approximation in Hilbert spaces \cite{Naylor1982_Book}. Further research is necessary to address these open questions and refine the proposed model.

\bibliographystyle{IEEEtran}
\bibliography{IEEEabrv, references}



\appendices


\section{Lemmas and corollaries}
Recall that we are employing $\mc{H} = \{\wh{\ul{H}}_i\}$ to denote a finite set $\{\wh{\ul{H}}_i,\; i = 1, \ldots, T\}$ of pairwise orthonormal Hermitian operators. 

\begin{lemma}
\label{lem:M_realsymmetric}
The QCM $\QCM{H}{v} = \{(\QCM{H}{v})_{ij}\}$ associated with the set of pairwise orthonormal Hermitian operators $\mc{H}$ is real and symmetric. 
\end{lemma}

\emph{Proof.} 
From Definition~\ref{def:QCM}, we have
\begin{align*}
  (\QCM{H}{v})_{ji}
    &= 0.5\, 
       \la \{\wh{\ul{H}}_j, \wh{\ul{H}}_i\} \ra_{\ul{v}} 
         - \la \wh{\ul{H}}_j \ra_{\ul{v}}
           \cdot
           \la \wh{\ul{H}}_i \ra_{\ul{v}} \\
    &= 0.5\, 
       \la \{\wh{\ul{H}}_i, \wh{\ul{H}}_j\} \ra_{\ul{v}} 
         - \la \wh{\ul{H}}_i \ra_{\ul{v}}
           \cdot
           \la \wh{\ul{H}}_j \ra_{\ul{v}}
       \notag \\
    &= (\QCM{H}{v})_{ij},
\end{align*}
because $\{\wh{\ul{H}}_i, \wh{\ul{H}}_j\} = \{\wh{\ul{H}}_j, \wh{\ul{H}}_i\}$ and both $\la \wh{\ul{H}}_i \ra_{\ul{v}}$ and $\la \wh{\ul{H}}_j \ra_{\ul{v}}$ are scalars. So, $\ul{M}$ is symmetric.

In addition, 
\begin{align*}
  (\QCM{H}{v})_{ij}^*
    &= 0.5\, \la \{\wh{\ul{H}}_i, \wh{\ul{H}}_j\} \ra_{\ul{v}}^* 
         - \la \wh{\ul{H}}_i \ra_{\ul{v}}^*
           \cdot
           \la \wh{\ul{H}}_j \ra_{\ul{v}}^* \\
    &= 0.5\, \la \ul{v} \mid \{\wh{\ul{H}}_i, \wh{\ul{H}}_j\} \mid \ul{v} \ra^* \\
    &\qquad\qquad
         - \la \ul{v} \mid \wh{\ul{H}}_i \mid \ul{v} \ra^*
           \cdot
           \la \ul{v} \mid \wh{\ul{H}}_j \mid \ul{v} \ra^* \\
    &= 0.5\, \la \ul{v} \mid \wh{\ul{H}}_i \wh{\ul{H}}_j \mid \ul{v} \ra^* 
         + 0.5\, \la \ul{v} \mid \wh{\ul{H}}_j \wh{\ul{H}}_i \mid \ul{v} \ra^* \\
    &\qquad\qquad
         - \la \ul{v} \mid \wh{\ul{H}}_i \mid \ul{v} \ra^*
           \cdot
           \la \ul{v} \mid \wh{\ul{H}}_j \mid \ul{v} \ra^* 
\end{align*}
But we note that 
\begin{alignat*}{2}
  &\la \ul{v} \mid \wh{\ul{H}}_i \wh{\ul{H}}_j \mid \ul{v} \ra^*
    &
      &= \la \wh{\ul{H}}_i \ul{v} \mid \wh{\ul{H}}_j \ul{v} \ra^* 
       = \la \wh{\ul{H}}_j \ul{v} \mid \wh{\ul{H}}_i \ul{v} \ra \\
  &
    &
      &= \la \ul{v} \mid \wh{\ul{H}}_j \wh{\ul{H}}_i \mid \ul{v} \ra; \\
  &\la \ul{v} \mid \wh{\ul{H}}_i \mid \ul{v} \ra^*
    &
      &= \la \ul{v} \mid \wh{\ul{H}}_i \ul{v} \ra^*
       = \la \wh{\ul{H}}_i \ul{v} \mid \ul{v} \ra \\
  &
    &
      &= \la \ul{v} \mid \wh{\ul{H}}_i \mid \ul{v} \ra.
\end{alignat*}
In a similar manner, we can also show that $\la \ul{v} \mid \wh{\ul{H}}_j \wh{\ul{H}}_i \mid \ul{v} \ra^* = \la \ul{v} \mid \wh{\ul{H}}_i \wh{\ul{H}}_j \mid \ul{v} \ra^*$ and $\la \ul{v} \mid \wh{\ul{H}}_j \mid \ul{v} \ra^* = \la \ul{v} \mid \wh{\ul{H}}_j \mid \ul{v} \ra$. When substituted into the expression above for $(\QCM{H}{v})_{ij}^*$, we see that $(\QCM{H}{v})_{ij}^* = (\QCM{H}{v})_{ij}$. So, $\ul{M}$ is real.
\QEDclosed

\begin{lemma}
\label{lem:Var_positive}
The variance $\tr{Var}\, (\wh{\ul{H}})_{\ul{\psi}} = \la \wh{\ul{H}}^2 \ra_{\ul{\psi}} - |\la \wh{\ul{H}} \ra_{\ul{\psi}}|^2$ of the Hermitian operator $\wh{\ul{H}}$ w.r.t. an arbitrary normalized vector $\ul{\psi},\; \la \ul{\psi} \mid \ul{\psi} \ra = 1$, satisfies
\[
  \tr{Var}\, (\wh{\ul{H}})_{\ul{\psi}}
    = \la \wh{\ul{H}}^2 \ra_{\ul{\psi}} - |\la \wh{\ul{H}} \ra_{\ul{\psi}}|^2 
    \geq 0;
\]
the equality holds iff $\ul{\psi}$ is an eigen-state of $\wh{\ul{H}}$.
\end{lemma}

\emph{Proof.}
For arbitrary $\alpha \in \mbb{C}$ and $\ul{\psi}$, we know that
\[
  0
    \leq
      \la \wh{\ul{H}}\, \ul{\psi} - \alpha\, \ul{\psi},\, \wh{\ul{H}}\, \ul{\psi} - \alpha\, \ul{\psi} \ra.
\]
Furthermore, the equality holds true iff $\wh{\ul{H}}\, \ul{\psi} - \alpha\, \ul{\psi} = 0$, i.e., iff $\ul{\psi}$ is an eigen-state of $\wh{\ul{H}}$. 

So, let us consider the case when $\ul{\psi}$ is not an eigen-state of $\wh{\ul{H}}$ so that 
\begin{align*}
  0
    &< \la \wh{\ul{H}}\, \ul{\psi} - \alpha\, \ul{\psi},\, \wh{\ul{H}}\, \ul{\psi} - \alpha\, \ul{\psi} \ra \\
    &= \la \wh{\ul{H}}\, \ul{\psi} \mid \wh{\ul{H}}\, \ul{\psi} \ra - \la \wh{\ul{H}}\, \ul{\psi} \mid \alpha\, \ul{\psi} \ra - \la \alpha\, \ul{\psi} \mid \wh{\ul{H}}\, \ul{\psi} \ra \\
    &\qquad\qquad
         + \la \alpha\, \ul{\psi} \mid \alpha\, \ul{\psi} \ra \\
    &= \la \wh{\ul{H}}\, \ul{\psi} \mid \wh{\ul{H}}\, \ul{\psi} \ra - \alpha\, \la \wh{\ul{H}}\, \ul{\psi}  \mid \ul{\psi} \ra - \alpha^* \la \ul{\psi} \mid \wh{\ul{H}}\, \ul{\psi} \ra \\
    &\qquad\qquad
         + |\alpha|^2\, \la \ul{\psi} \mid \ul{\psi} \ra.
\end{align*}
Since $\alpha$ is arbitrary, select
\[ 
  \alpha
    = \la \ul{\psi} \mid \wh{\ul{H}}\, \ul{\psi} \ra
  \implies
  \alpha^*
    = \la \wh{\ul{H}}\, \ul{\psi} \mid \ul{\psi} \ra.
\]
Substitute into the above inequality:
\begin{align*}
  0
    &< \la \ul{\psi} \mid \wh{\ul{H}}^2 \mid \ul{\psi} \ra - |\la \ul{\psi} \mid \wh{\ul{H}}\, \ul{\psi} \ra|^2 - |\la \ul{\psi} \mid \wh{\ul{H}}\, \ul{\psi} \ra|^2 \\
    &\qquad\qquad
    + |\la \ul{\psi} \mid \wh{\ul{H}}\, \ul{\psi} \ra|^2 \\
    &= \la \ul{\psi} \mid \wh{\ul{H}}^2 \mid \ul{\psi} \ra - |\la \ul{\psi} \mid \wh{\ul{H}} \mid \ul{\psi} \ra|^2
     = \la \wh{\ul{H}}^2 \ra_{\ul{\psi}} - |\la \wh{\ul{H}} \ra_{\ul{\psi}}|^2.
\end{align*}
This establishes the claim.
\QEDclosed

\begin{corollary}
\label{cor:H_eigen-state}
Given the Hermitian operator $\wh{\ul{H}}$ and the normalized vector $\ul{\psi},\; \la \ul{\psi} \mid \ul{\psi} \ra = 1$, $\tr{Var}\, (\wh{\ul{H}})_{\ul{\psi}} = 0$ iff $\ul{\psi}$, is an eigen-state  of $\wh{\ul{H}}$.
\end{corollary}

\begin{corollary}
\label{cor:M_NullSpace}
Suppose, for some arbitrary real-valued vector $\ul{w} \in \{w_i\} \in \mbb{R}^T$, $\wh{\ul{H}} = (\mc{H})_{\ul{w}}$. Then the following are true:
\begin{itemize}
  \item[(i)] For arbitrary $\ul{v} \in \mbb{R}^T$,  $\QCM{H}{v}$ is p.s.d. 
  \item[(ii)] $\ul{w} \in \NN{\QCM{H}{v}}$, the null space of the QCM $\QCM{H}{v}$ associated with $\mc{H}$ w.r.t $\ul{v}$ iff $\ul{v},\; \la \ul{v} \mid \ul{v} \ra = 1$, is an eigen-state of $\wh{\ul{H}}$. 
\end{itemize}
\end{corollary}

\emph{Proof.} 
\begin{itemize}
  \item[(i)] This follows directly from Lemma~\ref{lem:Var_positive} and \cite{Qi2019Quantum}.
  \item[(ii)] First, suppose $\ul{w} \in \NN{\QCM{H}{v}}$ so that 
\[
  \QCM{H}{v}\, \ul{w} 
    = 0
  \implies
  \la \QCM{H}{v} \ra_{\ul{w}}
    = \ul{w}^T \QCM{H}{v}\, \ul{w}
    = 0.
\]
Then, from \cite{Qi2019Quantum}, $\tr{Var}\, (\wh{\ul{H}})_{\ul{v}} = \la \QCM{H}{v} \ra_{\ul{w}} = 0$. Corollary~\ref{cor:H_eigen-state} then implies that $\ul{v}$ is an eigen-state of $\wh{\ul{H}}$.

Conversely, suppose $\la \QCM{H}{v} \ra_{\ul{w}} = \ul{w}^T \QCM{H}{v}\, \ul{w} = 0$. We know from Lemma~\ref{lem:M_realsymmetric} that $\QCM{H}{v}$ is real and symmetric. Let its SVD be
\[
  \QCM{H}{v}
    = \ul{U}\, \ul{\Sigma}\, \ul{U}^T,
\]
where $\ul{U} \in \mbb{R}^{T \times T}$ is unitary and $\ul{\Sigma} \in \mbb{R}^{T \times T}$ is diagonal with its diagonal entries being the non-negative singular values of $\QCM{H}{v}$. Let $\QCM{H}{v}^{1/2} = \ul{\Sigma}^{1/2} \ul{U}^T$. Then,  
\begin{align*}
  \ul{w}^T \QCM{H}{v}\, \ul{w}
    &= \ul{w}^T \QCM{H}{v}^{{1/2}^T}\, \QCM{H}{v}^{1/2} \ul{w} \\
    &= \la \QCM{H}{v}^{1/2} \ul{w}, \QCM{H}{v}^{1/2} \ul{w} \ra
     = 0,
\end{align*}
meaning that we must have 
\[
  \QCM{H}{v}^{1/2} \ul{w} 
    = 0 
  \implies 
  \QCM{H}{v}\, \ul{w} 
    = 0,
\]
i.e., $\ul{w} \in \NN{\QCM{H}{v}}$.
\QEDclosed
\end{itemize}


\end{document}